\documentclass[letterpaper]{article} 
\usepackage{aaai25}  
\usepackage{times}  
\usepackage{helvet}  
\usepackage{courier}  
\usepackage[hyphens]{url}  
\usepackage{graphicx} 
\usepackage{natbib}  
\usepackage{caption} 
\usepackage{graphicx}
\usepackage{booktabs}
\usepackage{url}
\usepackage{graphicx}
\usepackage{multirow}
\usepackage{multicol}
\usepackage{colortbl}
\usepackage{siunitx}
\usepackage{caption}
\usepackage{diagbox}
\usepackage{tabularx}
\usepackage{amsmath, amssymb}
\usepackage{makecell}
\usepackage{enumitem}
\usepackage{xspace}
\usepackage{subfigure}
\usepackage{etoc}
\usepackage{float}
\usepackage{xr}

\usepackage{newfloat}
\usepackage{listings}
\DeclareCaptionStyle{ruled}{labelfont=normalfont,labelsep=colon,strut=off} 
\floatstyle{ruled}
\newfloat{listing}{tb}{lst}{}
\floatname{listing}{Listing}
\title{Walking the Web of Concept-Class Relationships in Incrementally Trained Interpretable Models}
\author {
    Susmit Agrawal,
    Deepika Vemuri,
    Sri Siddarth Chakaravarthy P,
    Vineeth N. Balasubramanian
}
\affiliations {
    Indian Institute of Technology Hyderabad\\
    \{ai22mtech12002, ai22resch11001\}@iith.ac.in, \{sri.siddarth, vineethnb\}@cse.iith.ac.in
}

\begin{document}

\newcommand{\ours}[0]{MuCIL\xspace}
\newcommand{\oursbf}[0]{\textbf{MuCIL}\xspace}

\maketitle
\begin{abstract}
Concept-based methods have emerged as a promising direction to develop interpretable neural networks in standard supervised settings. However, most works that study them in incremental settings assume either a static concept set across all experiences or assume that each experience relies on a distinct set of concepts. In this work, we study concept-based models in a more realistic, dynamic setting where new classes may rely on older concepts in addition to introducing new concepts themselves. We show that concepts and classes form a complex web of relationships, which is susceptible to degradation and needs to be preserved and augmented across experiences. We introduce new metrics to show that existing concept-based models cannot preserve these relationships even when trained using methods to prevent catastrophic forgetting, since they cannot handle forgetting at concept, class, and concept-class relationship levels simultaneously. To address these issues, we propose a novel method - \oursbf - that uses multimodal concepts to perform classification without increasing the number of trainable parameters across experiences. The multimodal concepts are aligned to concepts provided in natural language, making them interpretable by design. Through extensive experimentation, we show that our approach obtains state-of-the-art classification performance compared to other concept-based models, achieving over 2$\times$ the classification performance in some cases. We also study the ability of our model to perform interventions on concepts, and show that it can localize visual concepts in input images, providing post-hoc interpretations.
\end{abstract}

\begin{links}
    \link{Code}{ https://github.com/Susmit-A/MuCIL}
    \link{Appendix}{https://susmit-a.github.io/misc/appendix.pdf}
\end{links}

\begin{figure}
\centering\includegraphics[width=0.85\linewidth]{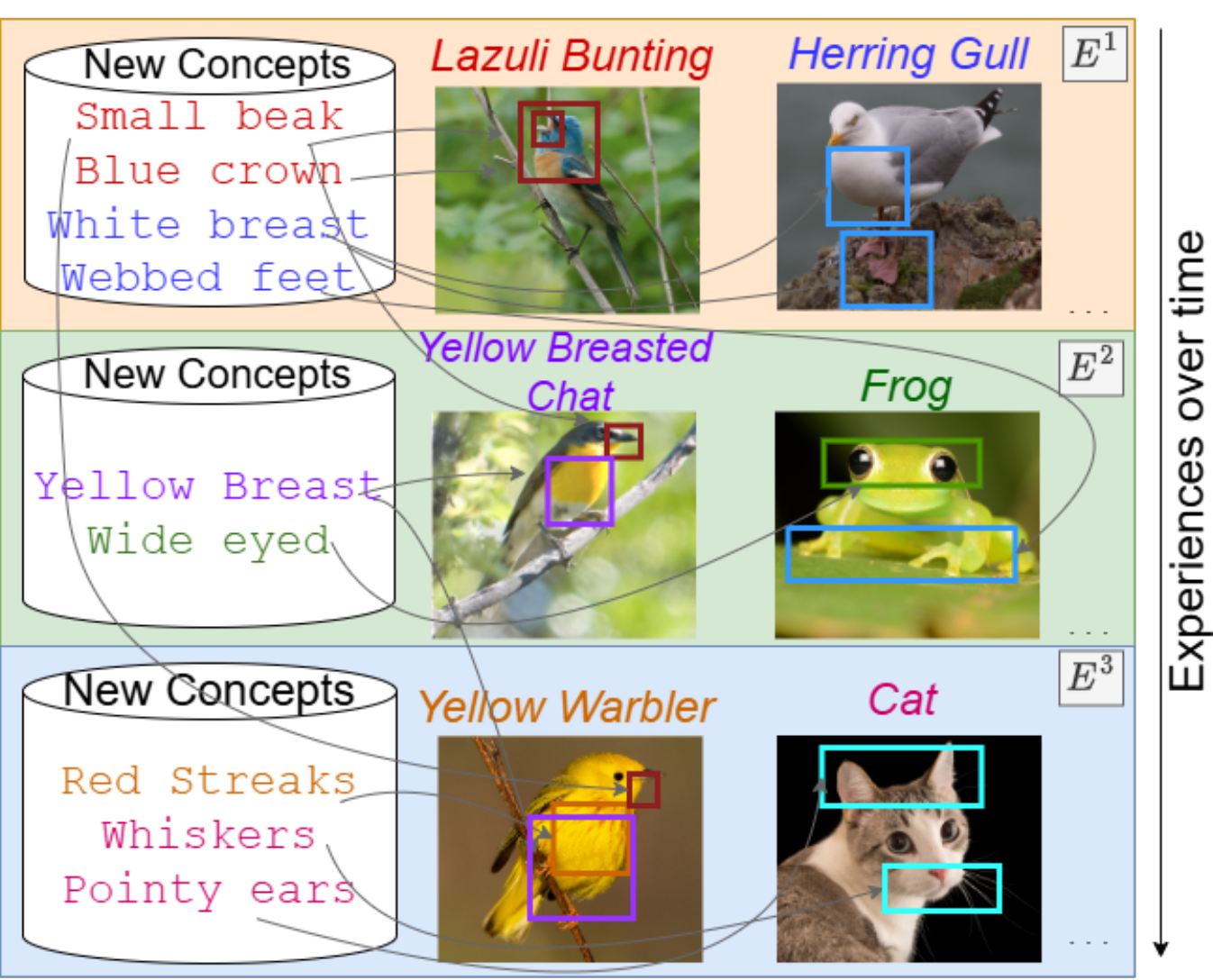}
    \caption{ Illustration of our setting. Concepts introduced in an experience are shared among classes from other experiences. \oursbf focuses on the difficult challenge of capturing and preserving this web of class-concept relationships over multiple experiences.}
    \label{fig:thumbnail}
\end{figure}

\section{Introduction}
Concept-based models have gained the attention of the computer vision community in recent years \citep{kim2023concept,margeloiu2021concept,barker2023selective,cbms}, as a means of interpreting the output prediction in terms of learned or human-defined atomic interpretable `\textit{concepts}'. 
These models attempt to predict the target class generally as a weighted linear combination of meaningful concepts.
For example, such a model may identify the concept set \textit{\{whiskers, four legs, pointy ears, sociable\}} as key semantic attributes that help make a prediction \textit{cat} on a given image. But with the recent growth of incremental model learning, a pertinent question arises -- can the abovementioned model be adapted later to identify \textit{golden retriever} in addition to \textit{cat}? The two classes share some common concepts such as \textit{\{four legs, sociable\}}, while having discriminative concepts such as \textit{\{whiskers, pointy ears\}} in case of \textit{cat} and \textit{\{golden fur, floppy ears\}} in case of \textit{golden retriever}. Incrementally learning such concept-based models with newer classes as well as concepts forms the key focus of this work.

Teaching neural network models about new distinctive concepts (e.g. \textit{golden fur}) of new classes (e.g. \textit{dog}), while reusing previously known concepts (\textit{four legs, sociable}) is a highly nontrivial problem. While concept-based models have seen a range of efforts in recent years, there has been very little work in incrementally learning such models. The limited existing efforts either assume that all new classes share the same pool of attributes as older ones \citep{marconato2022catastrophic}, or have independent non-overlapping attribute sets \citep{rymarczyk2023icicle}. In this work, we propose a more general and realistic setting in which classes seen at a later stage may share concepts from past classes while introducing new concepts of their own. This creates a complex web of concept-class relationships across experiences, as illustrated in Figure \ref{fig:thumbnail}, which needs to be preserved and expanded as the model learns to classify and explain new classes. Note that, as in standard incremental learning, the model is required to achieve good classification performance on newly introduced classes, while maintaining classification performance on previously seen ones.

While traditional deep learning models are susceptible to forgetting old classes as new ones are introduced (\textit{catastrophic forgetting} \citep{hadsell2020embracing}), we find that concept-based models are susceptible to forgetting \textit{concept-class relationships} as well. Overcoming two levels of forgetting in incrementally learning concept-based models presents new challenges that need to be addressed explicitly. 

To this end, we propose \ours, a novel \textbf{Mu}ltimodal \textbf{C}oncept-Based \textbf{I}ncremental \textbf{L}earner. 
We combine embeddings of text concepts, called \textit{concept anchors}, with image representations to create \textit{multimodal concept embeddings} for a given image. These embeddings are latent vectors containing information that helps in classification while also providing interpretations. 
We propose the use of \textit{concept grounding}, which allows the interpretation of multimodal concepts in terms of text-based concepts. All these components are integrated with the fundamental consideration that the model should be able to incorporate new classes and new concepts continually, while also forming new concept-class associations that may emerge in the process.

Our key contributions can be summarized as follows: (i) We propose a new method for the relatively new setting of concept-based incremental learning, where a model adapts dynamically to new classes as well as new concepts; (ii) We introduce \textit{multimodal concept embeddings}, a combination of image embeddings and interpretable concept anchors,  as part of our method to perform classification. Our approach is primarily intended to allow scalability of concept-based models to newer experiences without increase in parameters; (iii) We perform a comprehensive suite of experiments to evaluate our method on well-known benchmark datasets. We study our method's performance both in a incremental as well as standard supervised settings, achieving state of the art results; (iv) We propose three new metrics to evaluate concept-based models in the proposed setting: Concept-Class Relationship Forgetting, Concept Linear Accuracy and Active Concept Ratio. Our approach can offer concept-specific localizations implicitly as a means of interpreting the model prediction (see Figure \ref{fig:visual-grounding}), without being explicitly trained to do so.

\section{Related Work}
\label{sec_related_work}
\textbf{Interpretability of Deep Neural Network Models}: 
Interpretability methods in DNN models can be broadly classified into post-hoc and ante-hoc methods. Post-hoc methods aim to interpret model predictions after training \citep{8237336,chen2020adapting,8354201,sattarzadeh2021integrated,yvinec2022singe,benitez2023ante,sundararajan2020many,wang2021shapley,jethani2021fastshap,wang2021shapley,NIPS2017_0060ef47,fong2017interpretable,petsiuk2018rise,montavon2019explainable}. Recent efforts have highlighted the issues with post-hoc methods and their reliability in reflecting a model's reasoning \citep{rudin2019stop,vilone2021notions,nauta2023anecdotal}. On the other hand, ante-hoc methods that jointly learn to explain and predict provide models that are inherently interpretable \citep{sokol2021explainability,benitez2023ante}. Ante-hoc methods have also been found to provide interpretatations that help make the model more robust and reliable \citep{alvarezmelis2018robust, 9964439}. We focus on this genre of methods in this work. \cite{cbms} proposed Concept Bottleneck Models (CBMs), a method that uses interpretable, human-defined concepts, combining them linearly to perform classification. CBMs also allow human interventions on concept activations \citep{shin2023closer,steinmann2023learning} to steer the final prediction of a model. \citep{kim2023concept,collins2023human,LearningConcise} obtained the intermediate semantic concepts by replacing domain experts with Large Language Models (LLMs). This allows for ease and flexibility in obtaining the concept set.
Using LLMs to obtain concepts also allow grounding of neurons in a bottleneck layer to a human-understandable concept, an issue with CBMs that was highlighted in \citep{margeloiu2021concept}. Other concept-based methods \citep{alvarezmelis2018robust,Chen_2020,kazhdan2020cme,rigotti2021attention,benitez2023ante} use a different notion of concepts based on prototype representations (see appendix \S A1).

\noindent \textbf{Concept-Based Incremental Learning}:
While Incremental Learning in standard supervised settings has been widely explored \citep{wang2023comprehensive}, Concept-Based Incremental Learning has remained largely unstudied. We identify \citep{marconato2022catastrophic} as an early effort in this direction; however, this work trains CBMs in a continual setting under an assumption that all concepts, including those required for unseen classes, are accessible from the first experience itself, which does not emulate a real-world setting. More recently, \cite{rymarczyk2023icicle} proposed an interpretable CL method that uses part-based prototypes as concepts. As mentioned earlier, our notion of concepts allows us to go beyond parts of an object category, as in CBM-based models.

\section{\ours: Methodology}
\begin{figure*}[t]
\centering
\includegraphics[width=0.9\textwidth]{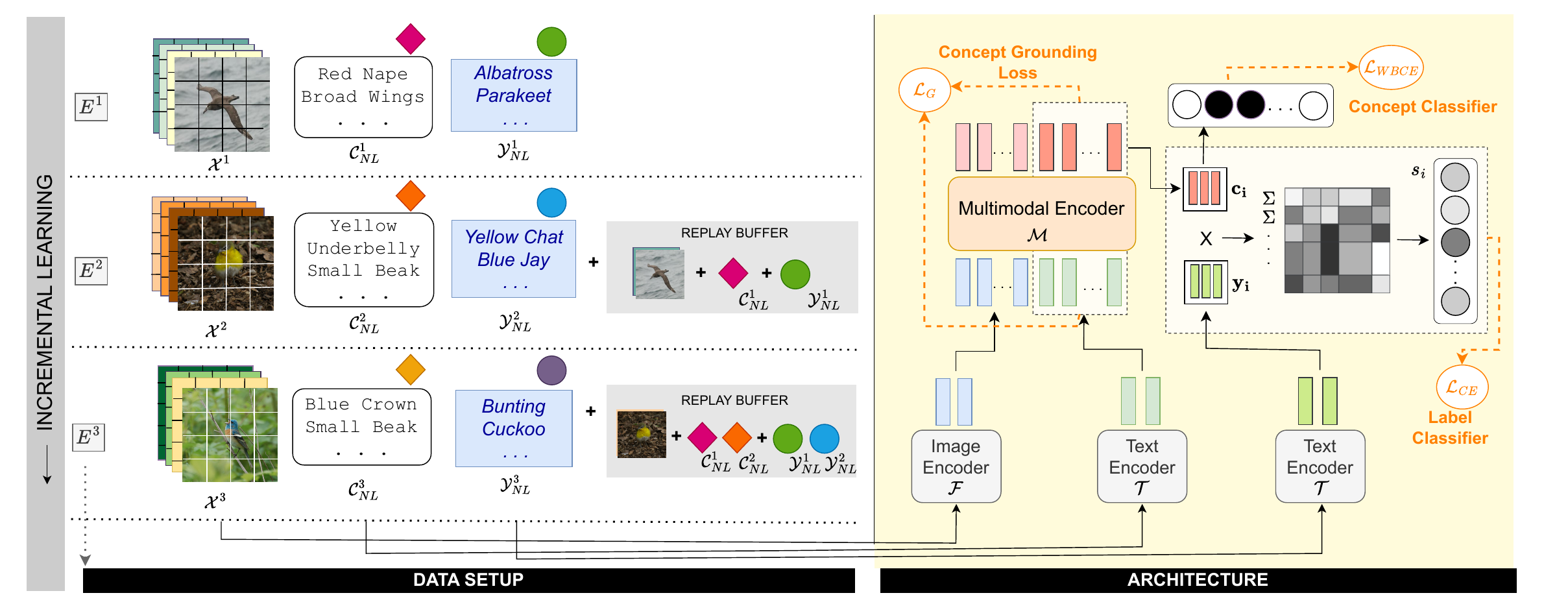}
    \caption{ \textbf{Overview of our setup and proposed architecture.} Our architecture receives new classes and associated concepts across multiple experiences in a CL setting. We use pre-trained language and vision encoders to get embeddings of the input image, concepts, and classes. These are then used to create multimodal image-concept embeddings using our \textit{Multimodal Encoder}. The multimodal concept embeddings are grounded to their concept anchors using a loss function and are used to predict both class labels and the presence/absence of corresponding concepts in the image.}
    \label{fig:overview}
\end{figure*}

\textbf{Preliminaries and Notations.}
Given a sequence of $T$ experiences, with each experience $i$ consisting of $n$ image-label pairs $(\mathcal{X}^i$, $\mathcal{Y}^i) = \{(x^i_1, y^i_1)$, $(x^i_2$, $y^i_2)$, ..., $(x^i_n, y^i_n)\}$, a class-incremental continual learning (CIL) system aims to learn an experience $t$ without catastrophically forgetting the previous ${t-1}$ experiences. In the scenario where finer class details are available as concepts for classification, each experience $i$ consists of $n$ image-label-concept tuples $(\mathcal{X}^i, \mathcal{Y}^i, \mathcal{C}^i) = \{(x^i_1, y^i_1, \mathcal{C}^i_1), \allowdisplaybreaks (x^i_2, y^i_2, \mathcal{C}^i_2), ..., (x^i_{n}, y^i_{n}, \mathcal{C}^i_{n})\}$, where $\mathcal{C}^i$ is the set of concepts known during experience $i$ and $\mathcal{C}^i_k$ is the set of active concepts in example $k$. The set of concepts known during $i$ is the union of all concept sets from experiences $1$ to $i$. We use the same active concept set for all instances of the same class, i.e., $\mathcal{C}^i_{k_1} = \mathcal{C}^i_{k_2}$ if $y^i_{k_1} = y^i_{k_2}$, as done in prior work \citep{cbms,LaBo,oikarinen2023labelfree}. Annotating instance-level concepts for large datasets is a challenging and error-prone task, requiring large amounts of time and effort by domain experts. Instead, using concepts at a class level allows us to specify the general characteristics of a class. We obtain these concepts directly from annotations in certain datasets (e.g. CUB) and use a Large Language Model (LLM) in other cases.
One could view this as a weakly supervised setting with noisy concept labels. 

\noindent \textbf{Challenges.} While concept-based models for image classification have grown in the community, extending them to learn across experiences and new classes incrementally is non-trivial, introducing new challenges. (1) \textbf{Forgetting at Two Levels:} In a traditional CIL setting, the catastrophic forgetting of previously learned classes in newer experiences is only at a class-level. In our proposed setting, it includes both concept-level and concept-class relationship forgetting. Beyond forgetting of concepts from previous experiences,
concept-class relationship forgetting can be more critical, affecting the model's ability to recognize previously encountered classes based on the concepts they were originally associated with. This affects the model's ability to understand and preserve how concepts relate to classes across experiences. (2) \textbf{Parameter Scaling:} Extending existing concept-based learning frameworks to a CIL setting would necessitate the addition of new learnable parameters with each new experience, leading to larger models, which is undesirable. Therefore, designing a method capable of integrating new classes and concepts without expanding the parameter space is ideal. (3) \textbf{Information Bottleneck:} Another challenge is the use of a single representation for encoding all relevant visual and concept information, which could lead to an information bottleneck as the system scales to accommodate more concepts and classes. (4) \textbf{Semantic Misrepresentations:} Lastly, some existing methods do not accurately capture the semantics of concepts \citep{margeloiu2021concept}. One widely used method to explicitly encode concept semantics is leveraging textual concept embeddings. These semantics should be respected and updated as new concepts arrive. A naive way to implement this is by projecting a network's output representation on pre-computed concept embeddings \citep{LaBo,oikarinen2023labelfree}, which amplifies the unified representation problem. We propose a more flexible method for overcoming this issue.

We propose a novel framework that addresses each of the abovementioned challenges. Our overall solution consists of multiple components:
(1) A \textbf{Multimodal Encoder}, that fuses visual information and textual concepts to create \textit{multimodal image-concept embeddings}. It is designed to handle an increasing number of concepts without growing the number of parameters. It also creates multiple distinct representations in parallel to avoid information bottlenecks while preserving concept semantics. (2) A \textbf{Parameter-free Clasifier} that uses the multimodal concept embeddings along with the textual descriptions of classes to perform classification. (3) \textbf{Concept Neurons} that predict whether or not a given concept exists in an image and also allow probing the model for interpretability.

\noindent \textbf{Multimodal Image-Concept Encoder.}
We introduce a novel multimodal image-concept encoder that can be integrated into any standard transformer architecture. It is used to merge image embeddings and textual concept embeddings, thereby generating a sequence of multimodal embeddings that capture both visual and concept inputs.

Our encoder incorporates new concepts without expanding its parameter space. This is made possible by the transformer's inherent ability to handle variable-length sequences, allowing the model to accommodate additional concepts by appending them to the input sequence without necessitating an increase in parameters. The use of transformer encoder layers enables the processing of mixed data inputs, resulting in a sequence of multimodal image and concept embeddings by simultaneously attending to the two modalities. Using a transformer is hence not a drop-in replacement to a different network architecture such as a CNN, but is a central part of our overall methodology.

Formally, the multimodal image-concept encoder, $\mathcal{M}$, is a stack of transformer encoder layers that receive image embeddings $\mathbf{x}_i$ as well as textual concept embeddings $\mathbf{c}_1 \mathbf{c}_2 ... \mathbf{c}_{|\mathcal{C}^{i}|}$ as input.
Note that this includes \textit{all concepts available until the current experience}.
The output of $\mathcal{M}$ is a sequence of vectors $\{\mathcal{X}'^{(i)},\mathcal{C}'^{(i)}\}$, where $\mathcal{X}'^{(i)}$ corresponding to the image patch embeddings $\{\mathbf{x}'_i\}$ and $\mathcal{C}'^{(i)} = \{\mathbf{c'_1, c'_2, ..., c'_{|\mathcal{C}^{i}|}}\}$ are concept embeddings. These embeddings combine both textual and visual information and hence are multimodal by design. Similar to standard transformer-based classification where all outputs except the CLS token are discarded, the $\mathcal{X}'^{(i)}$ plays no further part in our methodology.

The concept embeddings $\mathcal{C}'^{(i)}$ do not have any semantic grounding however; we hence use a \textit{Concept Grounding Loss} to map them to known natural language-based concept anchors, as below:

\begin{equation}
\label{eq:concept_grounding_loss}
\mathcal{L}_G = -\frac{1}{|\mathcal{C}^i|}\sum_{k=1}^{|\mathcal{C}^i|} \frac{\mathbf{c}_k \cdot (\mathbf{W}^T\mathbf{c}'_k + \mathbf{b})}{|\mathbf{c}_k| \cdot |\mathbf{W}^T\mathbf{c}'_k + \mathbf{b}|}
\end{equation}

\noindent $\mathbf{W}$ and $\mathbf{b}$ are learnable parameters that are shared among concepts. This loss semantically aligns the learned concept embeddings with their corresponding concept text anchors in the ground truth by ensuring that all original embeddings can be recovered using a single linear transform. Intuitively, this enforces each output concept vector to be associated with a deterministic representation of the concept, viz the text embedding. This deterministic association with a known ("grounded") reference guarantees that the vector has an associated semantic meaning, thus providing interpretability.

\noindent \textbf{Parameter-free incremental classification.} As stated earlier, existing concept-based learning frameworks use dense layers on top of concept embeddings, which makes it challenging to scale to newer experiences continually. We hence take inspiration from methods such as \citep{radford2021clip} to use text embeddings of classes to perform classification, thus allowing our method to be parameter-free in the classifier and allowing for scalability to newer experiences.

Classification is hence performed by returning the index of the class embedding which aligns most with the vectors in $\mathcal{C'}$ for a given input sample. The alignment between the $j$th concept embedding of the current sample, $\mathbf{c}'_j$, and the text embedding of the $k$th class $\mathbf{y}_k$ is computed as $\mathbf{c}'_j \cdot \mathbf{y}_k$, their dot product.
The final classification result is hence given by $\hat{y} = \operatorname*{argmax}_s (s_1, s_2, ..., s_{|\mathcal{Y}^i|})$, where $s_k = \sum_{j=1}^{|\mathcal{C}^i|} \mathbf{c}'_j \cdot \mathbf{y}_k$,
\noindent which returns the index of the class that aligns most strongly with concept embeddings of a given input. We use $s_k$ as the logit of class $k$, and perform a softmax operation on the logits to get classification probabilities, which are used to train the model with the standard cross-entropy loss $\mathcal{L}_{CE}$. We note that deriving class strengths from the concept embeddings in the above manner does not require additional parameters in newer experiences; this is a key element of our framework that enables scalability to both unseen classes and concepts when deployed in a CL setting.

\noindent \textbf{Concept Neurons.}
The presence of a concept is evaluated using a single shared dense layer with a sigmoid activation, denoted by $\sigma(\cdot)$, applied independently on each $\mathbf{c_i'} \in \mathcal{C}'$. A weighted binary cross-entropy loss $\mathcal{L}_{WBCE}$ is used to train the layer using provided ground-truth concept labels. We refer to the output logits of the layer $\sigma(\mathcal{C'})$ collectively as \textit{concept neurons}.
In \ours, concept neurons serve two purposes: (i) They provide additional supervision to learn concept embeddings better; and (ii) They provide an interface for identifying which concepts are active or inactive, thus enhancing interpretability. This allows for probing the model to evaluate the quality of concepts learned, and study how they change over experiences (see \S 4). We show a detailed illustration explaining them in appendix (\S A2).

\noindent \textbf{Details of $\mathcal{L}_{WBCE}$.}
For a given image, the ratio of the number of active concepts to the number of total concepts is quite small. This necessitates penalizing the misclassification of active concepts more strongly than the misclassification of inactive concepts. We do this by weighting the loss for active concepts by the fraction of \textit{inactive concepts}, and weighting the loss for inactive concepts by the fraction of \textit{active concepts}. The loss $\mathcal{L}_{WBCE}$ is then defined as:
{
\begin{align*}
    \mathcal{L}_{WBCE} &= \frac{\text{\# inactive concepts}}{\text{\# concepts}} \sum_{i=1}^{|\mathcal{C}^{\text{active}}|}\mathcal{L}_{BCE}(\sigma(\mathbf{c'}_i), 1) \\
    &+ \frac{\text{\# active concepts}}{\text{\# concepts}} \sum_{i=j}^{|\mathcal{C}^{\text{inactive}}|}\mathcal{L}_{BCE}(\sigma(\mathbf{c'}_j), 0)
\end{align*}
}

\noindent \textbf{Training Procedure.}
\ours thus has the following learnable components trained simultaneously end-to-end:  parameters of $\mathcal{M}$, parameters $\mathbf{W}$ and $\mathbf{b}$ in $\mathcal{L}_G$, and the weights of the dense layer used for obtaining concept neuron values, $\sigma(\cdot)$.
The global objective $\mathcal{L}$ is a weighted sum of the three different objectives:
\begin{equation}
    \mathcal{L} = \mathcal{L}_{CE} + \lambda_1 \mathcal{L}_{WBCE} + \lambda_2 \mathcal{L}_G
    \label{eqn:overall_loss}
\end{equation}

Intuitively, the terms $\mathcal{L}_{WBCE}$ and $\mathcal{L}_G$ control the concept-based learning in the framework. $\mathcal{L}_{WBCE}$ specifically trains $\mathcal{M}$ and $\sigma$ jointly to enable detection of the presence/absence of concepts, thus ensuring that visual features are properly represented in the multimodal concept vectors. $\mathcal{L}_{G}$ trains $\mathcal{M}, \mathbf{W},$ and $\mathbf{b}$ to enforce the concept embeddings to be grounded to their textual anchors. The $\mathcal{L}_{CE}$ term trains $\mathcal{M}$ to perform classification based on the outputs in $\mathcal{C'}$.

\noindent \textbf{Inference.} 
At test-time, given an image $x^{test}$ to be classified after experience $t$, we obtain its embedding vectors $\mathbf{x}^{test}$ and append them to the sequence of concept embeddings $\mathbf{c}_1 \mathbf{c}_2 ... \mathbf{c}_{|\mathcal{C}^{t}|}$. $\mathcal{M}$ receives this combined sequence as input and outputs a sequence of vectors $\{\mathcal{X'}^{(test)}, \mathcal{C'}^{(test)}\}$. $\mathcal{C'}^{(test)}$ is provided to the parameter-free classifier for obtaining the final class label. The active/inactive concepts in $x^{test}$ can be identified by thresholding the outputs of the concept neurons $\sigma(\mathcal{C'}^{(test)})$.

\section{Experiments and Results}
\label{sec_expts}

We perform a comprehensive suite of experiments to study the performance of \oursbf on well-known benchmarks: CIFAR-100, ImageNet-100 (INet-100), and CalTech-UCSD Birds 200 (CUB200). To study how our method works when concept annotations are provided with the dataset or otherwise, we use the human-annotated concepts provided in case of CUB200, and derive concepts for CIFAR-100 and ImageNet-100 from GPT 3.5 as described in \cite{oikarinen2023labelfree}. We study our method in standard supervised learning setting as well as in a Class-Incremental Learning setting as done in \cite{marconato2022catastrophic,rymarczyk2023icicle}, which we refer to as \textbf{SL} and \textbf{CL} settings respectively in our experiments. In the CL setting, we study our performance over 5 and 10 experiences using concept-based methods in conjunction with three well-known CL algorithms: Experience Replay (ER) \cite{rebuffi2017icarl}, A-GEM \cite{chaudhry2018efficient}, and DER++ \cite{buzzega2020derpp}, with a replay buffer size of 500 (we study other variations of buffer size in the Appendix). We select these CL algorithms since they can be adapted with concept-based model baselines. All implementation details (dataset details, architecture, hyperparameters) are in Appendix (\S A2,\S A4).

\noindent \textbf{Baselines.} 
We compare our method with existing works that perform concept-based class-incremental learning as well as by adapting other works that use concept-based learning to the CL setting (due to lack of many explicit efforts on this setting). Our baseline methods for comparison include: (i) ICIAP \citep{marconato2022catastrophic}, which 
makes the assumption that all concepts, including those that would likely only be provided in future experiences, are provided upfront;
(ii) \textit{Incremental CBM}, a version of the Concept Bottleneck Model \citep{cbms} that we modify to adapt to a class-incremental and concept-incremental learning scenario. We grow both the bottleneck layer and the linear classification layer as new classes and new concepts are introduced. 
We train these two baselines in sequential (-S) and joint (-J) settings as described in \cite{cbms}; (iii) We also compare with \textit{Label-Free CBM} \citep{oikarinen2023labelfree} and \textit{LaBo} \citep{LaBo}, variations of CBM that use projections of image embeddings onto natural language concept embeddings to form the bottleneck layer. Due to the frozen feature extractors and use of Generalized Linear Models (Label-Free) or Submodular Functions (LaBo) over the concept layer, extending these baselines to different CL algorithms is non-trivial. The same is applicable to CBM-S and ICIAP-S which involve multiple training stages with the feature extractor being fixed after the first stage. These methods are hence most compatible with ER, which we use for the corresponding baselines. Hence, for all ablation studies and analysis, we focus on using ER as the CL algorithm due to compatibility across all considered baseline models and to ensure fairness of comparison.

\begin{table*}[t]
    \centering
    \setlength{\tabcolsep}{2mm}
    \begin{tabular}{l c *{6}{S[table-format=1.4]} *{6}{S[table-format=1.4]} *{6}{S[table-format=1.4]}}
    \toprule
        & & \multicolumn{4}{c}{CIFAR-100} &  & \multicolumn{2}{c}{CUB} & \multicolumn{2}{c}{} & \multicolumn{2}{c}{INet-100} \\
    \cmidrule(lr){3-6} \cmidrule(lr){7-10} \cmidrule(lr){11-14}
    & & \multicolumn{2}{c}{5Exp} & \multicolumn{2}{c}{10Exp} & \multicolumn{2}{c}{5Exp} & \multicolumn{2}{c}{10Exp} &\multicolumn{2}{c}{5Exp} & \multicolumn{2}{c}{10Exp} \\
    \cmidrule(lr){3-4} \cmidrule(lr){5-6} \cmidrule(lr){7-8} \cmidrule(lr){9-10} \cmidrule(lr){11-12} \cmidrule(lr){13-14}
    \textbf{Method} & \textbf{CL Algo} & \multicolumn{1}{c}{FAA} & \multicolumn{1}{c}{AF} & \multicolumn{1}{c}{FAA} & \multicolumn{1}{c}{AF} & \multicolumn{1}{c}{FAA} & \multicolumn{1}{c}{AF} & \multicolumn{1}{c}{FAA} & \multicolumn{1}{c}{AF} & \multicolumn{1}{c}{FAA} & \multicolumn{1}{c}{AF} & \multicolumn{1}{c}{FAA} & \multicolumn{1}{c}{AF}\\
    \midrule
    
\textbf{CBM-J} \shortcite{cbms} & ER & \text{0.23} & \text{0.74} & \text{0.15} & \text{0.81} & \text{0.38} & \text{0.36} & \text{0.22} & \text{0.50} & \text{0.30} & \text{0.59} & \text{0.17} & \text{0.67} \\
\textbf{CBM-J} \shortcite{cbms} & A-GEM & \text{0.17} & \text{0.82} & \text{0.09} & \text{0.81} & \text{0.11} & \text{0.59} & \text{0.05} & \text{0.48} & \text{0.15} & \text{0.68} & \text{0.09} & \text{0.73} \\
\textbf{CBM-J} \shortcite{cbms} & DER++ & \text{0.22} & \text{0.76} & \text{0.13} & \text{0.84} & \text{0.38} & \text{0.40} & \text{0.27} & \text{0.49} & \text{0.27} & \text{0.64} & \text{0.16} & \text{0.69} \\

\textbf{ICIAP-J} \shortcite{marconato2022catastrophic} & ER & \text{0.25} & \text{0.72} & \text{0.13} & \text{0.82} & \text{0.38} & \text{0.36} & \text{0.22} & \text{0.50} & \text{0.31} & \text{0.57} & \text{0.18} & \text{0.67} \\
\textbf{ICIAP-J} \shortcite{marconato2022catastrophic} &  A-GEM & \text{0.17} & \text{0.79} & \text{0.09} & \text{0.83} & \text{0.11} & \text{0.59} & \text{0.05} & \text{0.48} & \text{0.17} & \text{0.67} & \text{0.09} & \text{0.74} \\
\textbf{ICIAP-J} \shortcite{marconato2022catastrophic} & DER++ & \text{0.22} & \text{0.76} & \text{0.14} & \text{0.83} & \text{0.38} & \text{0.40} & \text{0.27} & \text{0.49} & \text{0.28} & \text{0.63} & \text{0.15} & \text{0.70} \\

\midrule
\textbf{CBM-S} \shortcite{cbms} & ER & \text{0.30} & \text{0.69} & \text{0.28} & \text{0.71} & \text{0.35} & \text{0.46} & \text{0.22} & \text{0.54} & \text{0.29} & \text{0.61} & \text{0.21} & \text{0.65}  \\
\textbf{ICIAP-S} \shortcite{marconato2022catastrophic} & ER & \text{0.21} & \text{0.62} & \text{0.20} & \text{0.69} & \text{0.35} & \text{0.46} & \text{0.22} & \text{0.54} & \text{0.22} & \text{0.55} & \text{0.14} & \text{0.55} \\
\textbf{Label-Free} \shortcite{oikarinen2023labelfree} & ER & \text{0.22} & \text{0.34} & \text{0.19} & \textbf{0.24} & \text{0.31} & \text{0.38} & \text{0.42} & \text{0.48} & \text{0.07} & \text{0.31} & \text{0.11} & \text{0.26} \\
\textbf{LaBo} \shortcite{LaBo} & ER & \text{0.30} & \text{0.76} & \text{0.10} & \text{0.80} & \text{0.29} & \text{0.57} & \text{0.07} & \text{0.67} & \text{0.41} & \text{0.53} & \text{0.05} & \text{0.61} \\

    \midrule
    \textbf{\oursbf (Ours)} & ER & \text{0.67} & \text{0.35} & \textbf{0.63} & \text{0.38} & \text{0.78} & \text{0.11} & \textbf{0.76} & \text{0.14} & \text{0.80} & \text{0.09} & \text{0.79} & \textbf{0.09} \\
    
    \textbf{\oursbf (Ours)} & A-GEM & \text{0.36} & \text{0.73} & \text{0.44} & \text{0.59} & \text{0.30} & \text{0.71} & \text{0.15} & \text{0.82} & \text{0.66} & \text{0.27} & \text{0.71} & \text{0.19} \\

    \textbf{\oursbf (Ours)} & DER++ & \textbf{0.68} & \textbf{0.33} & \text{0.62} & \text{0.39} & \textbf{0.81} & \textbf{0.07} & \textbf{0.76} & \textbf{0.13} & \textbf{0.81} & \textbf{0.08} & \textbf{0.80} & \textbf{0.09} \\
    \bottomrule

    \end{tabular}
    \caption{CL performance (FAA = Final Average Accuracy, AF = Average Forgetting) of diff methods averaged over three random model initializations, on 5 and 10 experiences with buffer size 500, and three different CL algorithms. Top super-row contains baselines adapted to different CL methods. Middle super-row contains methods that are compatible with ER (adapting to other CL algorithms is non-trivial). Bottom super-row contains results for \oursbf trained using three different CL algorithms. \oursbf consistently delivers better performance than baselines (in all cases, std deviation $\leq$ 0.02). Additional results for ER with buffer sizes 2000 and 5000 have been provided in the Appendix (\S A3).}
    \label{tab:main_table}
\end{table*}

\noindent \textbf{Performance Metrics: Classification.} In the CL setting, we use two well-known performance metrics: \textit{Final Average Accuracy (FAA)} and \textit{Average Forgetting (AF)}. 
FAA is defined as: $FAA = \frac{1}{T}\sum_{i=1}^Tacc_i^T$, where $acc_i^T$ represents the model's accuracy on the validation split of experience $i$ after training on $T$ experiences. 
AF at experience $T$ is defined as: $AF = \frac{1}{T-1}\sum_{i=1}^{T-1} acc_i^i - acc_i^T$, i.e., the difference in accuracy on the validation set of experience $i$ when it was originally learned and the accuracy on it after the model has been trained on $T$ experiences. We use the standard \textit{Classification Accuracy} in the SL setting.

\noindent \textbf{New Performance Metrics: Concept Evaluation.} In order to evaluate the learned concepts and their evolution across experiences, we propose three new quantitive metrics: \textit{concept linear accuracy}, \textit{active concept ratio} and \textit{concept-class relationship forgetting}. We briefly describe each of these metrics, before presenting our results.

\noindent \underline{\textit{Concept Linear Accuracy:}} We use our concept neurons to evaluate how well concepts capture the relevant semantic information (for performing classification), and to study how they preserve this information over experiences.
The group of concept neurons are treated as a bottleneck layer, and a linear classifier is trained on top of the neuron logits. We denote the linear accuracy of concept neurons over a class set $\mathcal{Y}^i$ and a concept set $\mathcal{C}^j$ after training the model on $t$ experiences as $LA(t, \mathcal{Y}^i, \mathcal{C}^j)$, where $t$ is varied across CL experiences.

\noindent \underline{\textit{Concept-Class Relationship Forgetting:}} We define Concept-Class Relationship Forgetting (CCRF) of a concept set as the loss of its ability to provide relevant information to perform class-level discrimination over time. This can occur when concepts no longer align well to visual semantics in the provided image. This is different from forgetting in standard CL settings \cite{hadsell2020embracing} as the model may predict concepts correctly, but instead forgets how concepts correlate to classes. Mathematically, we measure CCRF as:
\begin{align}
    CCRF &= \frac{1}{T-1} \sum_{t=2}^{T} \frac{1}{t-1} \sum_{k=t-1}^{1} [LA(k, \mathcal{Y}^{k\setminus k-1}, \mathcal{C}^{k}) \notag \\
         &- LA(t, \mathcal{Y}^{k\setminus k-1}, \mathcal{C}^{k})]
    \label{eqn:ccrf}
\end{align}

\noindent The term $LA(k, \mathcal{Y}^{k\setminus k-1}, \mathcal{C}^{k}) - LA(t, \mathcal{Y}^{k\setminus k-1}, \mathcal{C}^{k})$ can also take negative values, indicating that the information captured by the concepts of experience $k$ is enhanced after training on the future experience $t$ instead of degrading. In our framework, the model generating concept logits for experience $i$ is the combination of $\mathcal{M}$ and $\sigma$ trained on experience $i$; in a CBM, this is the model upto the bottleneck layer.

\noindent \underline{\textit{Active Concept Ratio:}} To study the relevance of concepts to classes across experiences, we 
propose Active Concept Ratio (ACR) to measure how frequently concepts seen during experience $i$ activate when classifying images from experience $j$. A high value indicates that concepts from experience $i$ play an important role in understanding classes from experience $j$. Ideally, classes introduced in experience $i$ should have their highest ACR values associated with the concept set from the same experience, since those concepts best explain the classes. Positive ACR values with concept sets from other experiences indicate that those concepts are also activated in response to classes from experience $i$. Formally, let $N_j$ be the number of images to be classified in experience $j$. Then, the ACR for a concept set presented in experience $i$ for classifying images from experience $j$ is defined as $\left(\sum_{n=1}^{N_j}\hat{\mathcal{C}}^{i\setminus (i-1)}_n\right)/\left(\sum_{n=1}^{N_j}\hat{\mathcal{C}}^{i}_n\right)$. Here, $\hat{\mathcal{C}}_n^{i\setminus (i-1)}$ represents the model's (binary) predictions of \textit{unique concepts introduced} in experience $i$, while $\hat{\mathcal{C}}_n^{i}$ represents the model's (binary) predictions of \textit{all} concepts present in experience $i$.

\noindent \textbf{Results: CL Performance.} Table \ref{tab:main_table} shows our results on concept-based continual learning. 
Our approach outperforms all baselines, with significant margins on CIFAR-100 and ImageNet-100. 
This is done \textit{without adding any additional parameters} to our model with newer experiences, whereas other methods require new parameters to incorporate new classes and concepts. We also observe significantly lower forgetting across experiences.
These results show that our model readily incorporates knowledge about new concepts and classes while internally forming required concept-class associations, and also remembers these associations fairly well, when trained on new experiences. We also find that CL algorithms which explicitly replay labels (ER and DER++) perform better across all methods.

\noindent \textbf{Results: SL Performance.} To see how \ours fares in standard classification settings, 
we evaluate it in a full-data setting on the same three datasets. 
\begin{table}[ht]
\centering
    \begin{tabular}{lccc}
    \toprule
    \centering
    \textbf{Method} & \textbf{CIFAR-100} & \textbf{CUB} & \textbf{INet-100} \\
    \midrule
    \textbf{CBM-J} & 0.7868 & 0.7231 & 0.7773 \\
    \textbf{CBM-S} & 0.5712 & 0.6932 & 0.4265 \\
    \textbf{Label-Free} & 0.6431 & 0.7413 & 0.7818 \\
    \textbf{LaBo} & \textbf{0.8572} & 0.7015 & \textbf{0.8506} \\
    \oursbf \textbf{(Ours)} & \textbf{0.8567} & \textbf{0.8401} & 0.8466 \\
    \bottomrule
    \end{tabular}
    \caption{ Classification performance of different methods in the full-data (single experience) setting.}
    \label{tab:full-data}
\end{table}
We find that \ours considerably outperforms the next closest baseline on the CUB dataset, indicating that it is highly effective when used to differentiate between fine-grained classes. It also achieves comparable performance on ImageNet-100 and CIFAR-100, even though this setting is not our focus. 

\noindent \textbf{Results: CCRF}. We show the $CCRF$ metric values for \ours versus a jointly trained CBM in Table \ref{tab:ccrf}. We take the concept neurons (the bottleneck layer in the case of CBM-J) and train a linear layer on the different class set-concept set pairs required for the computation of the metric, as in Eqn \ref{eqn:ccrf}. We see that \ours is able to preserve significantly more information in its concept sets across experiences, as compared to CBM-J. Particulary on ImageNet100, we see an accuracy drop of over 14\% on average when training the classification layer on logits obtained from a bottleneck of a future experience. This indicates that CBM-J is unable to preserve concept-class relationships of a given experience after 
\begin{table}
    \begin{tabular}{lccc}
    \toprule
    \centering
    \textbf{Method} & \textbf{CIFAR-100} & \textbf{CUB} & \textbf{INet-100} \\
    \midrule
    \textbf{CBM-J} & 0.0929 & 0.0362 & 0.1450 \\
    \textbf{ICIAP-J} & 0.0941 & 0.0362 & 0.1632 \\
    \oursbf \textbf{(Ours)} & \textbf{0.0444} & \textbf{0.0099} & \textbf{0.0172} \\
    \bottomrule
    \end{tabular}
    \caption{ CCRF of CBM-J and \ours on benchmark datasets over 5 experiences with buffer size of 2000.}
    \label{tab:ccrf}
\end{table}
training on future experiences. In contrast, \ours only results in an accuracy drop of 1.7\%, indicating that concepts retain most of their information even after the model has been trained on future experiences. We find that ICIAP performs even worse, which we attribute to the fact that it uses the entire concept set in all experiences. This causes concepts of future experiences to first learn spurious semantics and later modify them. 

\begin{figure}[!h]
    \includegraphics[width=0.95\linewidth]{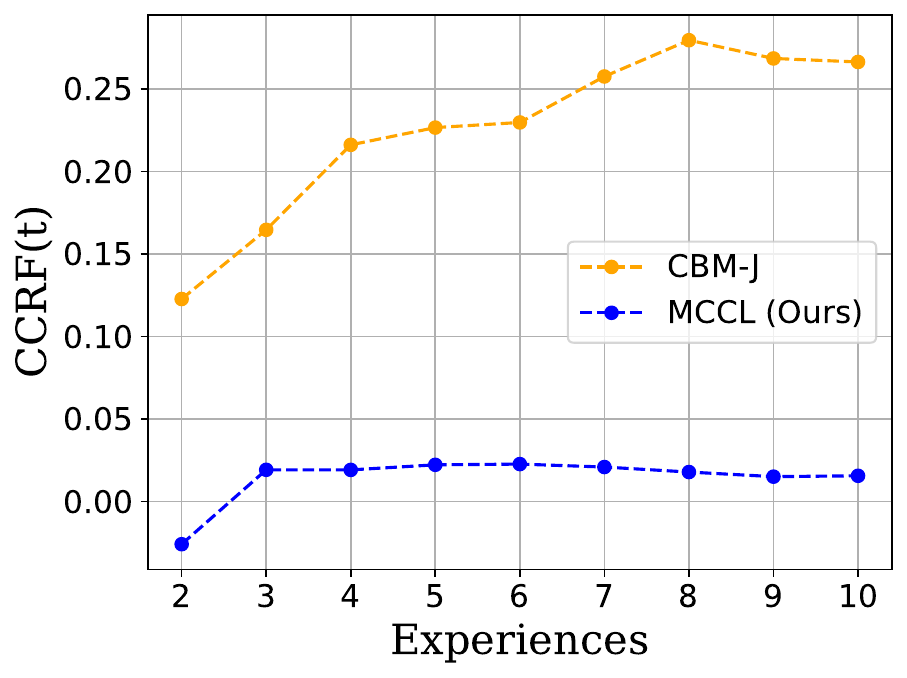}
    \caption{ Visualization of Concept-Class Relationship Forgetting on ImageNet-100 across 10 experiences, for CBM and \ours. While the relationship between concepts and classes deteriorates over new experiences, our method maintains forgetting at the same level.}
    \label{fig:ccrf_vis}
\end{figure}

\noindent \textit{Visualization of CCRF.} Table \ref{tab:ccrf} computed using Eqn \ref{eqn:ccrf} showcases the ability of \oursbf to preserve concept-class relationships. The same formulation can be used to also study experience-level CCRF. Fixing $t$ to some value gives us the form $CCRF(t) = \frac{1}{t-1} \sum_{k=t-1}^{1} LA(k, \mathcal{Y}^{k\setminus k-1}, \mathcal{C}^{k}) - LA(t, \mathcal{Y}^{k\setminus k-1}, \mathcal{C}^{k})$, 
which represents the average CCRF value for experience $t$ across previous concept set-class set pairs. This allows us to study how the concept-class relationships degrade over experiences. We show a visualization of this in Figure \ref{fig:ccrf_vis}. In the figure, we see that the relationships learned by the CBM model consistently degrade with experiences, while the relationships learned by our framework are preserved throughout.

\noindent \textbf{Results: ACR.} We present a visualization of ACRs across 10 experiences for a CBM, a Label-Free CBM and \ours model in
\begin{figure}
\centering
    \includegraphics[width=\linewidth]
{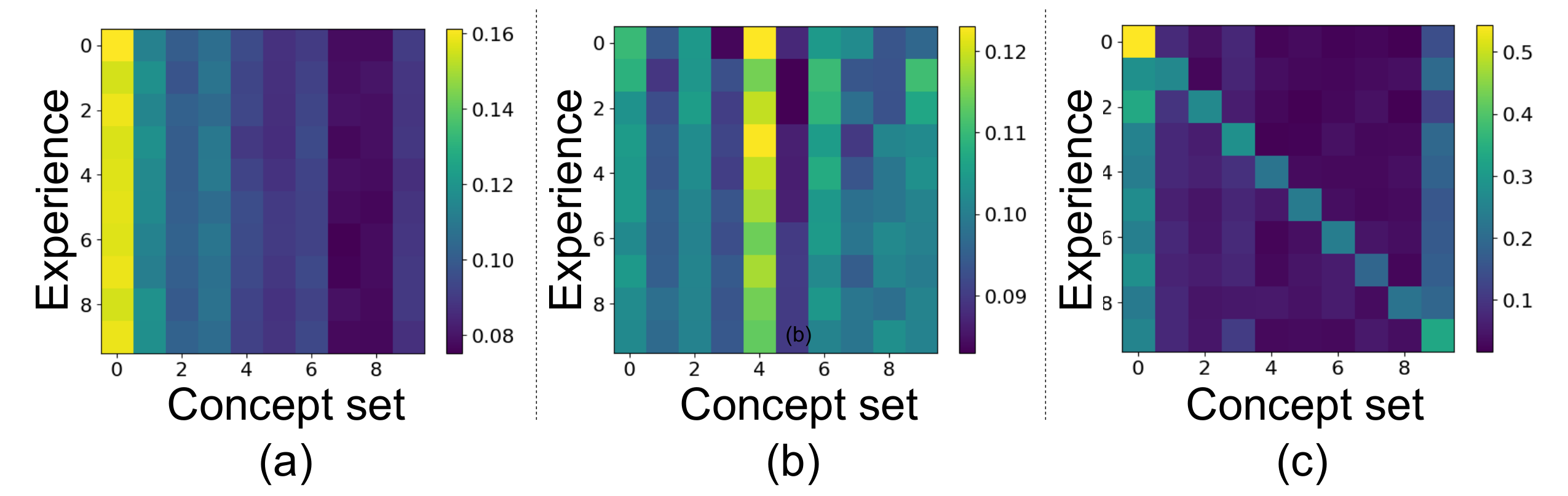}
        \caption{ Comparison of ACR matrices for (a) Standard CBM, (b) Label-Free CBM, and (c) \ours for CIFAR-100 over 10 experiences. Each row represents the ACR scores for the class sets of the corresponding experience, with each column representing the concept set of different experiences.}
        \label{fig:acr_vis}
\end{figure}
Fig. \ref{fig:acr_vis}. We see that the CBM is unable to effectively incorporate concepts that appear in later experiences and relies heavily on early concept sets. Label-Free CBM activates a given concept similarly across experiences, leading to poor explainability in terms of concepts required for a specific experience. 
\ours has a strong diagonal ACR matrix, showing that it strongly activates concepts that appear with (and therefore explain) a set of classes. It also appropriately activates earlier concepts, particularly fundamental concepts from experience one that are shared across future experiences. We see some bias toward the concept set introduced in the final experience, a common problem in most CL settings \cite{buzzega2021rethinking,Mai_2021_CVPR}. Addressing this can an interesting direction of future work. 

\noindent \textbf{Qualitative Results:}
\textbf{\textit{Visual Grounding and Attributions.}}
Our method can be directly used as a post-hoc analysis tool by computing visualizations of attention 
\begin{figure}[!h]
    \centering
    \includegraphics[width=\linewidth]{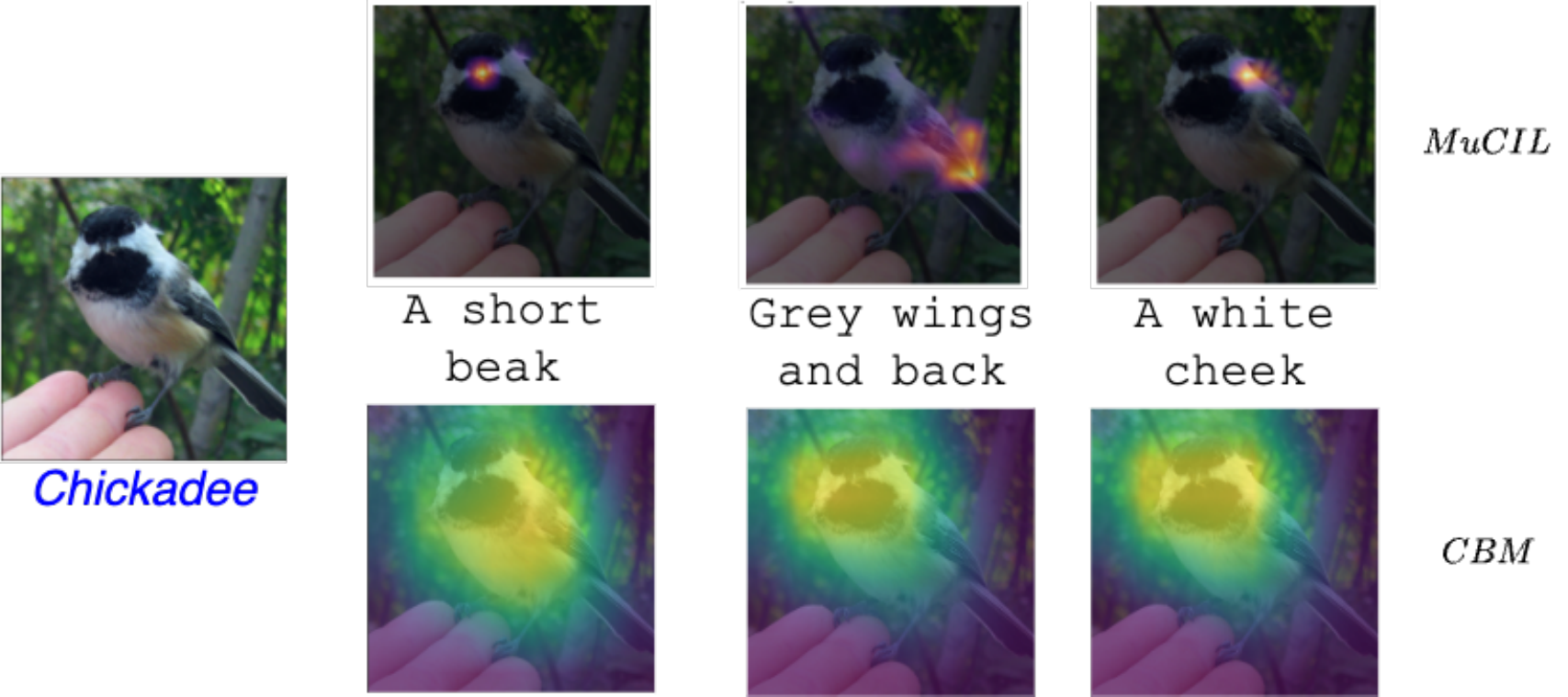}
     \caption{ \textbf{Visual grounding of part-based concepts:} Qualitative results for localizing visual concepts using \ours versus when localizing the same concepts using GradCAM on CBMs. More results provided in the Appendix (\S A3).}
    \label{fig:visual-grounding}
\end{figure}
maps learned by our model. Fig. \ref{fig:visual-grounding} shows one such result where our method has sharp focus on specific parts of the image corresponding to a concept. This supports the achievement of the objectives of our work.
In contrast, models without concept grounding have diffuse heatmaps without a clear focus on user-defined concepts \citep{margeloiu2021concept}. 

\noindent \textit{\textbf{Concept Replay Evaluation:}}
Results for $LA(T, \mathcal{Y}^T, \mathcal{C}^T)$, 
\begin{table}
\centering
    \begin{tabular}{lcccc}
    \toprule
    \textbf{Dataset} & \makecell{\textbf{FAA}\\w/ CR} & \makecell{\textbf{LA}\\w/ CR} & \makecell{\textbf{FAA}\\w/o CR} & \makecell{\textbf{LA}\\w/o CR} \\
    \midrule
    \textbf{CIFAR-100} & 0.7022 & 0.7650 & 0.6722 & 0.4511 \\
    \textbf{CUB200} & 0.8137 & 0.7914 & 0.7844 & 0.1382 \\
    \textbf{ImageNet-100} & 0.7970 & 0.7722 & 0.7903 & 0.5903 \\
    \bottomrule
    \end{tabular}
    \caption{ Linear layer training on top of concept neurons; CR = concept-augmented experience replay; LA = Accuracy of Linear Classifier trained on concept neurons.}
    \label{tab:concept-augmented-replay}
\end{table}
represented simply as $LA$, are shown in Table \ref{tab:concept-augmented-replay} with and without replaying concept labels from the buffer in addition to class labels (w/ CR and w/o CR respectively). 
Evidently, the concepts remain significantly more informative when using this enhanced Concept Replay.
CR also enhances overall performance in terms of FAA, as shown in the same table. We use CR with all benchmarked algorithms and baselines.

\noindent \textbf{\textit{Effects of $\mathcal{L}_{WBCE}$ \textit{and} $\mathcal{L}_{G}.$}:} We study the effects of these losses on performance by varying their weights $\lambda_1$ and $\lambda_2$ in Eqn \ref{eqn:overall_loss}. A low $\lambda_1$ results in a significant drop in performance in $LA$ scores. This indicates that the model forgets about relationships that exist between concept embeddings and classes, when $\mathcal{L}_{WBCE}$'s effect is reduced. Low values of $\lambda_2$ result in very low similarities between corresponding vectors in $\mathcal{C}^i$ and $\mathcal{C'}^{(i)}$, indicating that the concept embeddings lose semantic information encoded by their textual anchors 
when $\mathcal{L}_G$ has a reduced effect. We empirically found $\lambda_1 = 5$ and $\lambda_2 = 10$ to give the best performance overall in terms of FAA, LA and grounding similarity, with $LA = 0.7722$ and cosine similarity $0.998$. A detailed table showing the effects of these terms is in the Appendix (\S A3).

\noindent \textbf{\textit{Comparison with prototype-based methods}:} We compare our method's performance to ICICLE \cite{rymarczyk2023icicle} - a prototype-based approach for interpretable CIL - on CUB200 with 10 experiences. Our method significantly outperforms ICICLE -- we achieve an FAA score of 0.7606 \textit{without} experience IDs using 500 exemplars, whereas ICICLE achieves an FAA score of 0.602 \textit{with} experience IDs, and a score of 0.185 without experience IDs. We note that the two methods are different fundamentally in terms of the methodology; we provide the numbers here for completeness of understanding of our method's performance.

\begin{figure}[!h]
    \centering
    \includegraphics[width=\linewidth]{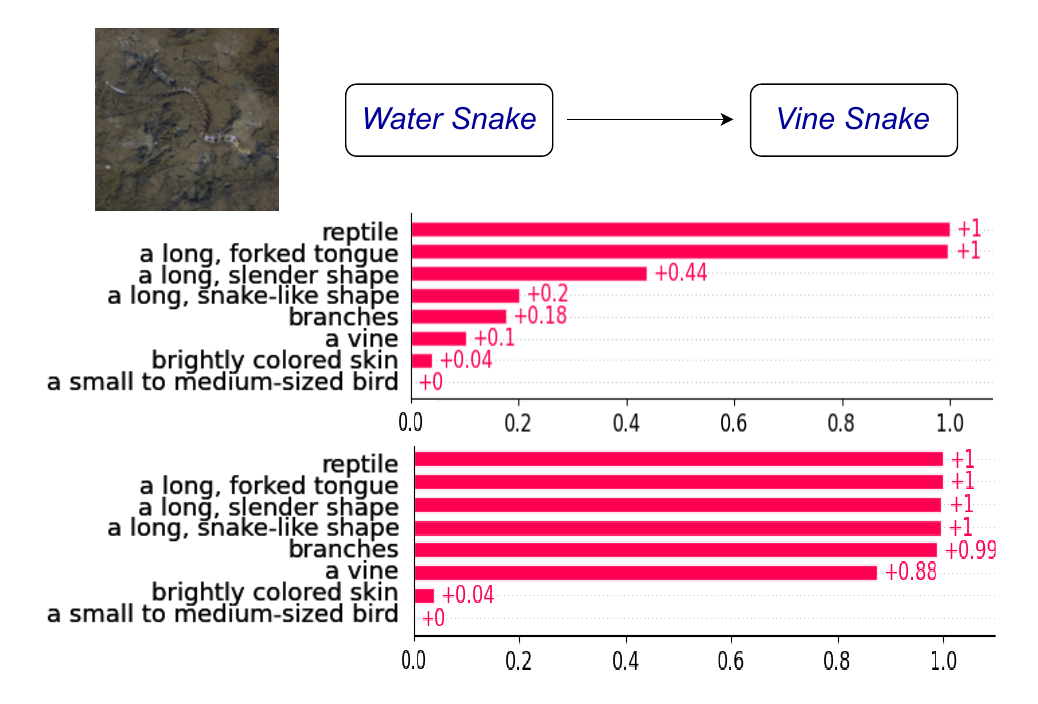}
    \caption{\footnotesize{Manual interventions on concepts: We identify concepts that are incorrectly labeled, and modify them based on the image semantics, this results in correct classification.}}
    \label{fig:interventions}
\end{figure}

\noindent \textbf{\textit{Evaluating concepts using interventions:}} 
To study the goodness of the learned concepts, we consider samples that are misclassified by a linear layer trained on top of concept neurons and perform interventions on the wrongly predicted concepts using the mechanism in \citep{cbms}. Performing interventions on a few key misclassified concepts converts the classification to a correct class label. 
This highlights the goodness of semantics of the learned concepts, and its impact on classification. Qualitative results for interventions are provided in Figure \ref{fig:interventions} and the Appendix (\S A3).

\noindent \textbf{\textit{Alignment between Similar Concepts.}} Ideally, if two concepts are similar in nature (and thus have similar text embeddings), their multimodal embeddings should maintain this similarity irrespective of whether or not the concepts co-occur in a given input. To verify if such inter-concept semantics are preserved, we compute pairwise concept embedding similarities for INet100 and the pairwise multimodal embedding similarities averaged over test samples. The elementwise squared errors between these matrices have a mean of 0.019, indicating that two similar concepts would likely have similar multimodal embeddings. This shows that our approach preserves relative semantics of concepts.

\noindent \textbf{\textit{Evaluation with Other Variants of Attention.}}
Recent attempts \citep{katharopoulos_et_al_2020,vyas_et_al_2020,shen2020efficient,wang2020linformer,kitaev2020reformer} have been made to improve the computational efficiency of transformers. To study the ability of our framework in incorporating new advancements in the transformer architecture, we evaluated a variant of our method that implemented \oursbf with Linear Attention.

\noindent We use transformer blocks featuring linear attention proposed in \citep{katharopoulos_et_al_2020} in our multimodal encoder. We evaluate this variant of our model across benchmark datasets for 5 experiences with 2000 exemplars. The results in Table \ref{tab:method_linear} show that even with linear attention, it achieves
comparable performance to vanilla attention on all three benchmarks. This could help make our method even more efficient, while retaining its performance. We believe this could be an interesting direction of future work.
 \begin{table}[h]
     \centering
     \setlength{\tabcolsep}{1mm}
     \begin{tabular}{@{}l|@{\hskip 2pt}ccccccc@{}}
     \toprule
         & \multicolumn{2}{c}{CIFAR-100} & \multicolumn{2}{c}{CUB} & \multicolumn{2}{c}{ImageNet-100} \\
     \cmidrule(lr){2-3} \cmidrule(lr){4-5} \cmidrule(lr){6-7}
     \textbf{Method} & {FAA} & {AF} & {FAA} & {AF} & {FAA} & {AF} \\
     \midrule
     \oursbf & \textbf{0.70} & \textbf{0.30} & \textbf{0.81} & 0.06 & \textbf{0.80} & 0.09 \\
     \oursbf\textbf{-L} & 0.69 & 0.31 & \textbf{0.82} & \textbf{0.05} & \textbf{0.80} & \textbf{0.08} \\
    \bottomrule
     \end{tabular}
     
     \caption{ Performance comparison of \ours vs. \ours-Linear over 5 experiences with 2000 exemplars, across different datasets.}
     \label{tab:method_linear}
\end{table}

\section{Conclusions and Future Work}
In this work, we study a new paradigm of continual learning for interpretable models where both new classes and new concepts are introduced across experiences. We show that existing works suffer from degradation in concept-class relationships in such settings. We propose a method that adapts the transformer architecture in vision-language encoders to include concept embeddings, which are anchored to natural language concepts.
Through a set of carefully designed loss terms, our approach can not only classify reliably in a CL setting, but can also specify the human-understandable concepts used for the classification.
We evaluate our method on three benchmark datasets and introduce new metrics to study the efficacy of concepts in our framework. Our qualitative and quantitative results show the significant promise of the proposed method. 
Beyond being a method for concept-based CL, we believe that our efforts can further open up the direction of inherently interpretable CL models in the community. Analysis of CL methods that do not directly replay past labels and the refinement of our architecture for better CL would be interesting directions of future work. 

\section*{Acknowledgments}
This work was partly supported by the Reliance Postgraduate Scholarship, Prime Minister’s Research Fellowship (PMRF) program and funding support from Adobe. We are grateful to the anonymous reviewers for their valuable feedback, which improved the presentation of the paper.

\bibliography{aaai25}

\clearpage

\newpage

\section*{Appendix}
\renewcommand{\thesection}{A}
\renewcommand{\thefigure}{A\arabic{figure}}
\renewcommand{\thetable}{A\arabic{table}}

In this appendix, we include the following details which we could not include in the main paper owing to space constraints. We will release our code publicly on acceptance, and provide all other implementation details herein.

\section*{Table of Contents}
\addcontentsline{toc}{section}{Table of Contents}
\begin{enumerate}
    \item[A.1] {Concepts: What and Why?} \dotfill 1
    \item[A.2] {Architecture and Implementation Details} \dotfill 1
    \item[A.3] {More Results and Analysis} \dotfill 3
    \item[A.4] {Dataset Descriptions} \dotfill 4
    \item[A.5] {Limitations, Future Directions and Broader Impact} \dotfill 5
    \item[A.6] {Per-Experience Performance Results} \dotfill 6
\end{enumerate}

\subsection{Concepts: What and Why?} \label{subsec:concepts_what_why}

In this work, we refer to \textit{intermediate text semantics that describe a class label} as ``concepts''. We state this to explicitly differentiate from other uses of the term. In recent literature in the field, concepts have been interpreted as:
(1) (Soft) binary labels that may not directly be grounded on human-understandable semantics and attempt to reconstruct a class label, as in \citep{cbms}; (2) Visual features that commonly appear in most instances of a given class, as in  \citep{rymarczyk2023icicle}; and (3) intermediate text semantics that describe a class label, as in \citep{oikarinen2023labelfree, LaBo}. We follow the third characterization in this work. The first connotation has been observed to not capture intended semantics \citep{margeloiu2021concept}, while the second conntation of prototypes are effective in certain settings, like instance-specific explanations. Our approach follows recent work \citep{oikarinen2023labelfree, LaBo} in viewing concepts as a compelling means to learn via explanations, rather than learn via prediction. Below are a few advantages of our approach:

\begin{itemize}[leftmargin=*]
    \item \textit{Generalizable Abstraction for Intermediate Semantics:} Text attributes provide a means to capture intermediate semantics that represents what a model is `thinking' or `considering important'. Unlike prototypes, which are based on specific instances or examples from training data, such text attributes can be generalized across instances. This abstraction facilitates a more human-relatable understanding of the underlying relationships that a model has learned.

    \item \textit{Human-Interpretable:} Concepts, as used herein, provide a means to connect latent embeddings inside a transformer to human-interpretable text, providing a pathway to better understand a model's functioning. Prototypes, while illustrative, may not provide this degree of interpretability, especially when the prototypes are derived from complex or non-intuitive examples. Text attributes can serve to succinctly communicate what features or aspects in the data influence the model's decisions.
    
    \item \textit{Flexibility and Adaptability:} Text attributes offer flexibility in adapting to different models and contexts. They can be easily modified, combined, or expanded upon to suit the specific needs of an experience or to improve interpretability. This adaptability is particularly beneficial in complex domains where the model's functioning needs to be thoroughly understood and communicated. With Large Language Models, they are also feasible to obtain at a class label level rather inexpensively.

    \item \textit{Avoidance of Overfitting to Specific Instances:} Relying on prototypes can sometimes lead to overfitting to specific instances in the training data, which may not generalize well to new, unseen data. Text attributes, by focusing on general concepts rather than specific examples, promote a more generalizable understanding of the model's behavior.

    \item \textit{Low Cost:} With Large Language Models, it is very feasible to obtain text attributes/concepts at a class label level at relatively low cost. (Note that we only need concepts at a class level, and not at a instance level.)

    \item \textit{Integration with Explanation Frameworks:} While this is the not the explicit focus of our work herein, text attributes can be integrated with existing explanation frameworks, such as feature importance measures, decision trees, or rule-based explanations. Such integration can enhance the comprehension and utility of explainability tools, making them more actionable for users. 

    \item \textit{Other Use Cases:} Text attributes can also be used to address other use cases such as biases in AI models by highlighting sensitive or critical concepts that require scrutiny. This is more challenging with prototypes, as the selection of representative examples may inadvertently reinforce existing biases or overlook subtle but important biases in the model's decisions on one particular sample.
    
\end{itemize}

\subsection{Architecture and Implementation Details}\label{subsec:architecture_implementationdetails}

\begin{figure}
    \centering

\includegraphics[width=0.95\linewidth]{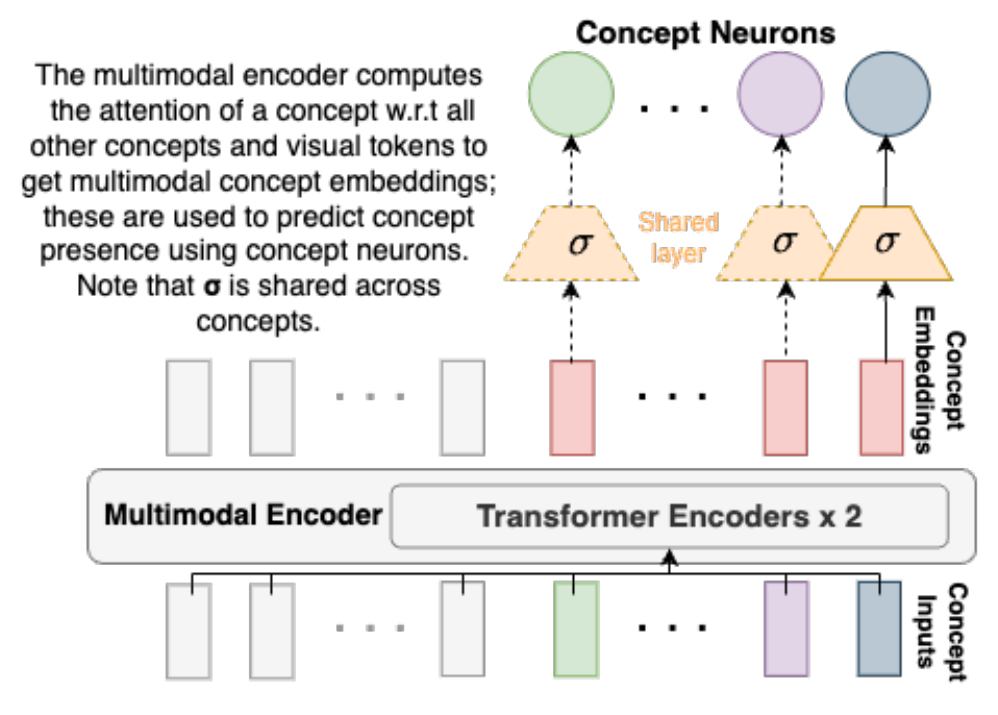}
        \caption{Concept Embeddings and Concept Neurons}
        \label{fig:multimodal_enc}
\end{figure}

\noindent \textit{Illustration of Concept Neurons.}
In Fig \ref{fig:multimodal_enc}, we provide an illustration of concept neurons. As shown in the figure, the same layer is applied on top of all multimodal embeddings to obtain the neuron values.

\noindent \textit{Use of VLMs.} In order to study if Vision-Language Model (VLM) pre-alignment is necessary for our method, we empirically study 9 different VL encoder pairs: CLIP \citep{radford2021clip} and FLAVA \citep{singh2022flava}, which have pre-aligned vision-language encoders, as well as BERT \citep{devlin2019bert} and ViT \citep{dosovitskiy2021vit} models trained on unimodal data, where our model explicitly aligns the modalities.

\begin{table}
\centering

\begin{tabular}{@{}l|@{\hskip 2pt}cccc@{}}
\toprule
\textbf{Dataset} & \multicolumn{1}{c}{\diagbox[width=6em]{\textbf{Vision}}{\textbf{Text}}} &  \textbf{FLAVA} & \textbf{CLIP} & \textbf{BERT} \\
\midrule
 & \textbf{FLAVA} & 0.625 & 0.625 & 0.627 \\
CIFAR-100 & \textbf{CLIP} & 0.642 & 0.631 & 0.621 \\
& \textbf{ViT} & \textbf{0.712} & 0.635 & 0.660 \\
\midrule
 & \textbf{FLAVA} & 0.709 & 0.642 & 0.667 \\
ImageNet-100 & \textbf{CLIP} & 0.735 & 0.693 & 0.708 \\
 & \textbf{ViT} & \textbf{0.792} & 0.758 & 0.737 \\
\midrule
 & \textbf{FLAVA} & 0.722 & 0.553 & 0.677 \\
CUB200 & \textbf{CLIP} & 0.756 & 0.670 & 0.717 \\
& \textbf{ViT} & \textbf{0.822} & 0.780 & 0.807 \\ 
\bottomrule
\end{tabular}

\caption{ FAA on benchmark datasets for diff VLMs}
\label{tab:VLalign}
\end{table}
Table \ref{tab:VLalign} presents the results which indicate that using pre-aligned VL models is not optimal; our method's performance using dedicated encoders is, in fact, superior to pre-aligned models.
We hypothesize this is because pre-aligned VL models are trained at a general image level, while our explicit approach allows more fine-grained association between image and text.

\noindent \textit{Vision, Language, and Multimodal Encoders ($\mathcal{F}, \mathcal{T}, \mathcal{M}$).} We use the FLAVA \citep{singh2022flava} language encoder paired with the ViT \citep{dosovitskiy2021vit} image encoder. Each encoder has a latent embedding dimension of 768. Our multimodal encoder uses two stacked transformer blocks with the same latent embedding size. In case any encoder does not have the required dimension (e.g. CLIP), we add a single trainable linear layer to project the embeddings to the required dimensions. We use the \textit{HuggingFace}\footnote{https://huggingface.co/} library to implement the transformer in case of the Full Attention version, or the \textit{fast-transformers}\footnote{https://fast-transformers.github.io/} library to implement transformer blocks with Linear Attention. The experimental setup is implemented using \textit{PyTorch}\footnote{https://pytorch.org/}, and all experiments are run on a single RTX 3090 GPU.

\noindent \textit{Training Hyperparameters.}
For Cifar100, we train the model for 10 epochs in every experience, with a starting learning rate of 0.001 and a batch size of 48. In the case of CUB, we train our model for 25 epochs in every experience and stop training after 15 epochs if the model converges. We start with a learning rate of 0.0003 and a batch size of 64. In the case of Imagenet100, we train the model for 5 epochs in every experience, with a starting learning rate of 0.001 and batch size of 48. In all three cases, we use Cosine Annealing to schedule the learning rate, decreasing it down to 0.0001.

\noindent \textit{Additional Details about Concept Bottlenecks.} A Concept Bottleneck layer, introduced by \cite{cbms}, is a layer where each neuron corresponds to a specific concept. Models containing such bottlenecks can be trained sequentially or jointly with the classification layer. Sequential and joint settings are applicable when the model contains a bottleneck layer followed by a classification layer. In the sequential setting, the model is first trained to predict concept labels. Post-training, a classifier is trained on top of concept logits predicted for the input images. The model and classifier are optimized separately. In joint training, both concept predictions and the classifier are trained end-to-end and optimized jointly.

\noindent \textit{Attention Head Visualization}
We build upon known methods of attention head visualization to extract image regions that the model focuses on for a particular concept. Briefly, our algorithm consists of three steps, shown in Fig \ref{fig:appendix-attnvis}: 
\begin{figure}[ht]
  \begin{center}
\includegraphics[width=0.55\linewidth]{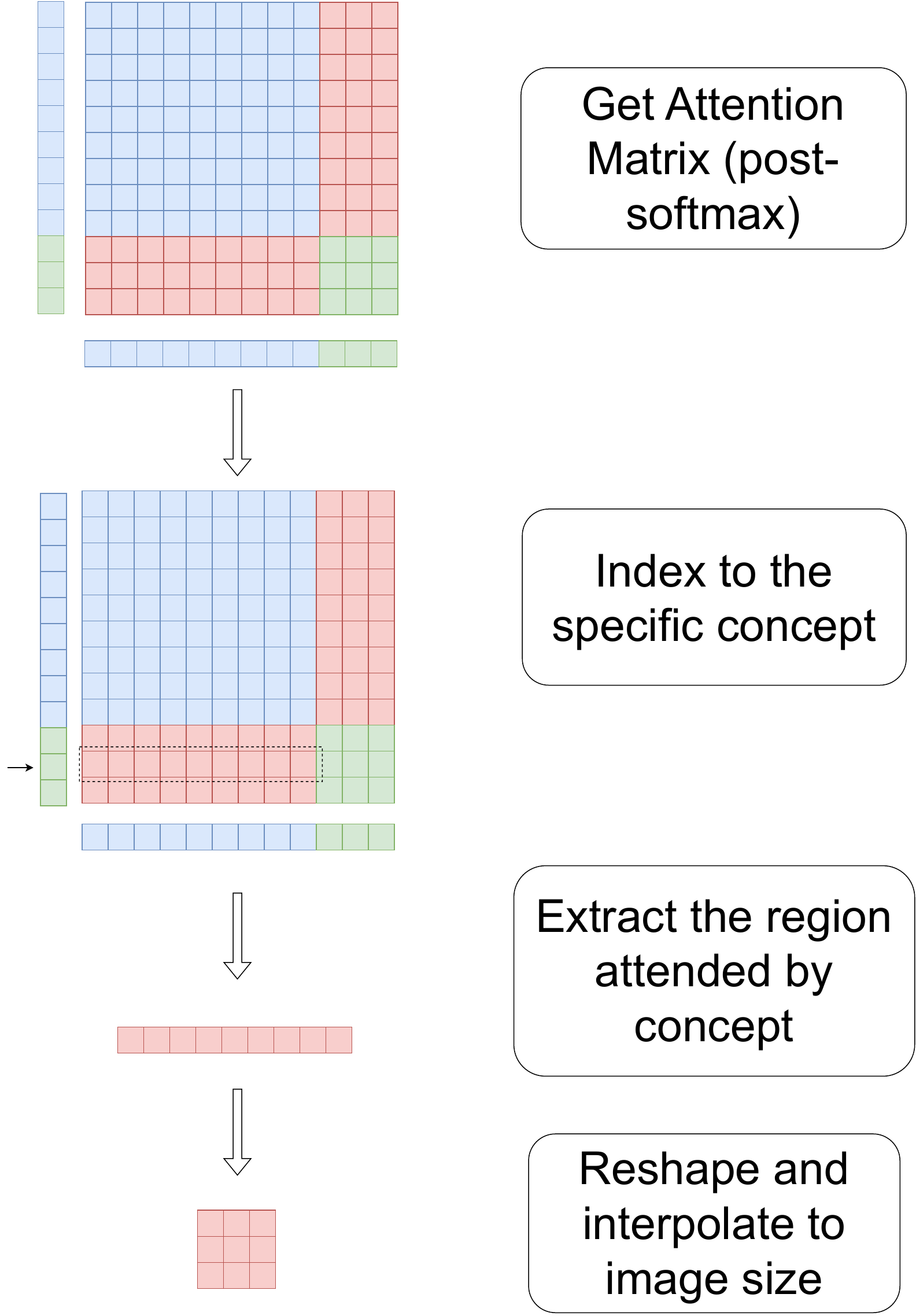}
    \caption{ Our attention visualization method. We induce localization via indexing through concepts instead of through CLS tokens, as done by other methods.}
    \label{fig:appendix-attnvis}
    \end{center}
\end{figure}
(1) We extract the post-softmax attention matrix of a chosen head after an input image and concept set has been processed. (2) The row indexed by the concept to be visualized gives attention scores over all concepts and image patches. Since we are only interested in the image patches, we extract the corresponding subcolumns from the concept row to get a single vector. The size of this vector is same as the number of visual tokens. (3) Post extraction, we visualize the attended regions by reshaping the extracted vector and performing interpolation to get a heatmap of the same size as the image.

\subsection{More Results and Analysis} \label{subsec:appendix_analysis}
\textit{Varying Buffer Sizes for Experience Replay:} We compare our method with baseline methods for different buffer sizes of 500, 2000, and 5000 on 5 and 10 experiences. The results for a buffer size of 500 were provided in the main paper, in table \ref{tab:main_table}. In table \ref{tab:main_table_appendix_2k} and \ref{tab:main_table_appendix_5k}, we show the results for other buffer sizes using ER.

\begin{table*}[t]
    \centering
    \begin{tabular}{l c *{5}{S[table-format=1.4]} *{6}{S[table-format=1.4]} *{6}{S[table-format=1.4]}}
    \toprule
        & \multicolumn{4}{c}{CIFAR-100} & & \multicolumn{2}{c}{CUB} & \multicolumn{2}{c}{} & \multicolumn{2}{c}{ImageNet-100} \\
    \cmidrule(lr){2-4} \cmidrule(lr){6-9} \cmidrule(lr){10-13}
    & \multicolumn{2}{c}{5Exp} & \multicolumn{2}{c}{10Exp} & \multicolumn{2}{c}{5Exp} & \multicolumn{2}{c}{10Exp} &\multicolumn{2}{c}{5Exp} & \multicolumn{2}{c}{10Exp} \\
    \cmidrule(lr){2-3} \cmidrule(lr){4-5} \cmidrule(lr){6-7} \cmidrule(lr){8-9} \cmidrule(lr){10-11} \cmidrule(lr){12-13}
    \textbf{Method} & \multicolumn{1}{c}{FAA} & \multicolumn{1}{c}{AF} & \multicolumn{1}{c}{FAA} & \multicolumn{1}{c}{AF} & \multicolumn{1}{c}{FAA} & \multicolumn{1}{c}{AF} & \multicolumn{1}{c}{FAA} & \multicolumn{1}{c}{AF} & \multicolumn{1}{c}{FAA} & \multicolumn{1}{c}{AF} & \multicolumn{1}{c}{FAA} & \multicolumn{1}{c}{AF}\\
    \midrule
    \textbf{CBM-S} \shortcite{cbms} & \text{0.47} & \text{0.47} & \text{0.49} & \text{0.47} & \text{0.53} & \text{0.20} & \text{0.46} & \text{0.26} & \text{0.51} & \text{0.35} & \text{0.47} & \text{0.38} \\
    \textbf{CBM-J} \shortcite{cbms} & \text{0.39} & \text{0.55} & \text{0.34} & \text{0.59} & \text{0.56} & \text{0.11} & \text{0.46} & \text{0.20} & \text{0.46} & \text{0.40} & \text{0.39} & \text{0.44} \\
    \textbf{ICIAP-S} \shortcite{marconato2022catastrophic} & \text{0.39} & \text{0.42} & \text{0.37} & \text{0.49} & \text{0.53} & \text{0.20} & \text{0.46} & \text{0.26} & \text{0.41} & \text{0.33} & \text{0.36} & \text{0.37} \\
    \textbf{ICIAP-J} \shortcite{marconato2022catastrophic} & \text{0.39} & \text{0.53} & \text{0.32} & \text{0.61} & \text{0.56} & \text{0.11} & \text{0.46} & \text{0.20} & \text{0.47} & \text{0.39} & \text{0.37} & \text{0.46} \\
    \textbf{Label-Free} \shortcite{oikarinen2023labelfree} & \text{0.34} & \textbf{0.24} & \text{0.30} & \textbf{0.19} & \text{0.45} & \text{0.45} & \text{0.58} & \text{0.47} & \text{0.09} & \text{0.30} & \text{0.17} & \text{0.30} \\
    \textbf{LaBo} \shortcite{LaBo} & \text{0.30} & \text{0.70} & \text{0.10} & \text{0.79} & \text{0.31} & \text{0.50} & \text{0.07} & \text{0.61} & \text{0.44} & \text{0.48} & \text{0.06} & \text{0.56} \\
    \midrule
    \textbf{\oursbf (Ours)} & \textbf{0.71} & \text{0.29} & \textbf{0.67} & \text{0.33} & \textbf{0.82} & \textbf{0.06} & \textbf{0.81} & \textbf{0.07} & \textbf{0.79} & \textbf{0.09} & \textbf{0.80} & \textbf{0.08} \\
    \bottomrule
    \end{tabular}
    \caption{ Continual learning performance of different methods averaged over three random model initializations, on 5 and 10 experiences with buffer size 2000. Our method delivers consistently better performance than the baselines.}
    \label{tab:main_table_appendix_2k}
\end{table*}

\begin{table*}[t]
    \centering
    \begin{tabular}{l c *{5}{S[table-format=1.4]} *{6}{S[table-format=1.4]} *{6}{S[table-format=1.4]}}
    \toprule
        & \multicolumn{4}{c}{CIFAR-100} & & \multicolumn{2}{c}{CUB} & \multicolumn{2}{c}{} & \multicolumn{2}{c}{ImageNet-100} \\
    \cmidrule(lr){2-4} \cmidrule(lr){6-9} \cmidrule(lr){10-13}
    & \multicolumn{2}{c}{5Exp} & \multicolumn{2}{c}{10Exp} & \multicolumn{2}{c}{5Exp} & \multicolumn{2}{c}{10Exp} &\multicolumn{2}{c}{5Exp} & \multicolumn{2}{c}{10Exp} \\
    \cmidrule(lr){2-3} \cmidrule(lr){4-5} \cmidrule(lr){6-7} \cmidrule(lr){8-9} \cmidrule(lr){10-11} \cmidrule(lr){12-13}
    \textbf{Method} & \multicolumn{1}{c}{FAA} & \multicolumn{1}{c}{AF} & \multicolumn{1}{c}{FAA} & \multicolumn{1}{c}{AF} & \multicolumn{1}{c}{FAA} & \multicolumn{1}{c}{AF} & \multicolumn{1}{c}{FAA} & \multicolumn{1}{c}{AF} & \multicolumn{1}{c}{FAA} & \multicolumn{1}{c}{AF} & \multicolumn{1}{c}{FAA} & \multicolumn{1}{c}{AF}\\
    \midrule
    \textbf{CBM-S} \shortcite{cbms} & \text{0.57} & \text{0.33} & \text{0.54} & \text{0.39} & \text{0.59} & \text{0.11} & \text{0.52} & \text{0.19} & \text{0.60} & \text{0.24} & \text{0.58} & \text{0.27} \\
    \textbf{CBM-J} \shortcite{cbms} & \text{0.50} & \text{0.38} & \text{0.45} & \text{0.44} & \text{0.57} & \text{0.06} & \text{0.51} & \text{0.12} & \text{0.53} & \text{0.31} & \text{0.49} & \text{0.34} \\
    \textbf{ICIAP-S} \shortcite{marconato2022catastrophic} & \text{0.46} & \text{0.32} & \text{0.47} & \text{0.39} & \text{0.59} & \text{0.11} & \text{0.52} & \text{0.19} & \text{0.52} & \text{0.23} & \text{0.46} & \text{0.28} \\
    \textbf{ICIAP-J} \shortcite{marconato2022catastrophic} & \text{0.52} & \text{0.36} & \text{0.47} & \text{0.42} & \text{0.57} & \text{0.06} & \text{0.51} & \text{0.12} & \text{0.55} & \text{0.29} & \text{0.49} & \text{0.33} \\
    \textbf{Label-Free} \shortcite{oikarinen2023labelfree} & \text{0.40} & \textbf{0.21} & \text{0.33} & \textbf{0.18} & \text{0.47} & \text{0.46} & \text{0.62} & \text{0.35} & \text{0.18} & \text{0.19} & \text{0.35} & \text{0.18} \\    
    \textbf{LaBo} \shortcite{LaBo} & \text{0.31} & \text{0.65} & \text{0.10} & \text{0.77} & \text{0.31} & \text{0.46} & \text{0.08} & \text{0.59} & \text{0.43} & \text{0.46} & \text{0.06} & \text{0.54} \\
    \midrule
    \textbf{\oursbf (Ours)} & \textbf{0.74} & \text{0.25} & \textbf{0.71} & \text{0.29} & \textbf{0.83} & \textbf{0.03} & \textbf{0.82} & \textbf{0.06} & \textbf{0.81} & \textbf{0.08} & \textbf{0.80} & \textbf{0.08} \\
    \bottomrule
    \end{tabular}
    \caption{ Continual learning performance of different methods averaged over three random model initializations, on 5 and 10 experiences with buffer size 5000. Our method delivers consistently better performance than the baselines.}
    \label{tab:main_table_appendix_5k}
\end{table*}

\textit{Sensitivity to $\lambda_1$ and $\lambda_2$:} We evaluate these hyperparameters by performing a grid search over different values. The results show that they directly relate to the interpretability aspects of the model, as described in the main paper (shown in Table \ref{tab:lamda1_2-faa}). We see that FAA values are within statistical error of the values of Table 1 in the main paper, implying that $\lambda$ values do not significantly affect classification accuracy. We show the alignment scores ($-\mathcal{L}_G$) in Table \ref{tab:LG-lamda} and Linear Accuracy scores in Tables \ref{tab:lambda-af}, and see a much wider range, indicating the impact of $\lambda$ values on these metrics.

\begin{table}[ht]
    \centering
    \setlength{\tabcolsep}{1mm}
    \begin{tabular}{@{}l|@{\hskip 2pt}ccccccc@{}}
        \toprule
        \textbf{Dataset} & \multicolumn{1}{c}{\diagbox[width=3em]{\textbf{$\lambda_1$}}{\textbf{$\lambda_2$}}} &  \textbf{0} & \textbf{0.001} & \textbf{1} & \textbf{5} & \textbf{10} & \textbf{100}\\
        \midrule
        & \textbf{0} & \text{0.80} & \text{0.79} & \text{0.80} & \text{0.81} & \text{0.81} & \text{0.79} \\
        & \textbf{0.001} & \text{0.81} & \text{0.81} & \text{0.80} & \text{0.81} & \text{0.81} & \text{0.79} \\
        CUB200 & \textbf{1} & \text{0.80} & \text{0.81} & \text{0.80} & \text{0.82} & \text{0.82} & \text{0.78} \\
        & \textbf{5} & \text{0.81} & \text{0.81} & \text{0.81} & \text{0.81} & \text{0.82} & \text{0.78} \\
        & \textbf{10} & \text{0.82} & \text{0.81} & \text{0.82} & \text{0.83} & \text{0.82} & \text{0.79} \\
        & \textbf{100} & \text{0.81} & \text{0.80} & \text{0.81} & \text{0.81} & \text{0.81} & \text{0.80} \\
        \midrule
        & \textbf{0} & \text{0.80} & \text{0.79} & \text{0.78} & \text{0.79} & \text{0.80} & \text{0.79} \\
        & \textbf{0.001} & \text{0.79} & \text{0.79} & \text{0.79} & \text{0.80} & \text{0.81} & \text{0.80}\\
        INet100 & \textbf{1} & \text{0.79} & \text{0.78} & \text{0.79} & \text{0.80} & \text{0.81} & \text{0.79} \\
        & \textbf{5} & \text{0.81} & \text{0.79} & \text{0.79} & \text{0.79} & \text{0.80} & \text{0.79} \\
        & \textbf{10} & \text{0.81} & \text{0.80} & \text{0.80} & \text{0.80} & \text{0.80} & \text{0.80} \\
        & \textbf{100} & \text{0.79} & \text{0.80} & \text{0.78} & \text{0.78} & \text{0.79} & \text{0.79} \\
        \bottomrule
    \end{tabular}
    
    \caption{ FAA on ImageNet-100 and CUB datasets for different values of $\lambda_1$ and $\lambda_2$.}
    \label{tab:lamda1_2-faa}
\end{table}

\begin{table}[ht]
    \centering
    \setlength{\tabcolsep}{1mm}
    \begin{tabular}{@{}l|@{\hskip 2pt}ccccccc@{}}
        \toprule
        \textbf{Dataset} & \multicolumn{1}{c}{\diagbox[width=3em]{\textbf{$\lambda_1$}}{\textbf{$\lambda_2$}}} &  \textbf{0} & \textbf{0.001} & \textbf{1} & \textbf{5} & \textbf{10} & \textbf{100}\\
        \midrule
        & \textbf{0} & \text{4e-32} & \text{0.79} & \text{0.94} & \text{0.98} & \text{0.99} & \text{1.00} \\
        & \textbf{0.001} & \text{8e-32} & \text{0.80} & \text{0.94} & \text{0.98} & \text{0.99} & \text{1.00} \\
        CUB & \textbf{1} & \text{1e-31} & \text{0.82} & \text{0.95} & \text{0.98} & \text{0.99} & \text{1.00} \\
        & \textbf{5} & \text{1e-31} & \text{0.83} & \text{0.94} & \text{0.98} & \text{0.99} & \text{1.00} \\
        & \textbf{10} & \text{1e-31} & \text{0.84} & \text{0.94} & \text{0.97} & \text{0.98} & \text{1.00} \\
        & \textbf{100} & \text{1e-31} & \text{0.86} & \text{0.90} & \text{0.94} & \text{0.95} & \text{0.99} \\
        \midrule
        & \textbf{0} & \text{9e-33} & \text{0.38} & \text{1.00} & \text{1.00} & \text{1.00} & \text{1.00} \\
        & \textbf{0.001} & \text{7e-32} & \text{0.41} & \text{1.00} & \text{1.00} & \text{1.00} & \text{1.00} \\
        INet100 & \textbf{1} & \text{1e-31} & \text{0.53} & \text{1.00} & \text{1.00} & \text{1.00} & \text{1.00} \\
        & \textbf{5} & \text{2e-31} & \text{0.69} & \text{0.99} & \text{1.00} & \text{1.00} & \text{1.00} \\
        & \textbf{10} & \text{7e-32} & \text{0.71} & \text{0.99} & \text{1.00} & \text{1.00} & \text{1.00} \\
        & \textbf{100} & \text{3e-31} & \text{0.72} & \text{0.90} & \text{0.98} & \text{0.99} & \text{1.00} \\
        \bottomrule
    \end{tabular}
    
    \caption{ $-\mathcal{L}_G$ on ImageNet-100 and CUB datasets for different values of $\lambda_1$ and $\lambda_2$.}
    \label{tab:LG-lamda}
\end{table}

\begin{table}[ht]
\centering
\setlength{\tabcolsep}{1mm}
\begin{tabular}{@{}l|@{\hskip 2pt}ccccccc@{}}
\toprule
\textbf{Dataset} & \multicolumn{1}{c}{\diagbox[width=3em]{\textbf{$\lambda_1$}}{\textbf{$\lambda_2$}}} &  \textbf{0} & \textbf{0.001} & \textbf{1} & \textbf{5} & \textbf{10} & \textbf{100}\\
\midrule
& \textbf{0} & \text{0.01} & \text{0.01} & \text{0.01} & \text{0.01} & \text{0.01} & \text{0.01} \\
& \textbf{0.001} & \text{0.01} & \text{0.02} & \text{0.02} & \text{0.01} & \text{0.02} & \text{0.01} \\
CUB & \textbf{1} & \text{0.75} & \text{0.75} & \text{0.75} & \text{0.79} & \text{0.80} & \text{0.80} \\
& \textbf{5} & \text{0.77} & \text{0.76} & \text{0.77} & \text{0.80} & \text{0.79} & \text{0.79} \\
& \textbf{10} & \text{0.78} & \text{0.78} & \text{0.79} & \text{0.79} & \text{0.79} & \text{0.80} \\
& \textbf{100} & \text{0.78} & \text{0.77} & \text{0.79} & \text{0.79} & \text{0.80} & \text{0.80} \\
\midrule
& \textbf{0} & \text{0.01} & \text{0.01} & \text{0.01} & \text{0.01} & \text{0.01} & \text{0.01} \\
& \textbf{0.001} & \text{0.03} & \text{0.01} & \text{0.12} & \text{0.13} & \text{0.04} & \text{0.03} \\
INet100 & \textbf{1} & \text{0.76} & \text{0.80} & \text{0.78} & \text{0.78} & \text{0.78} & \text{0.78} \\
& \textbf{5} & \text{0.80} & \text{0.79} & \text{0.80} & \text{0.80} & \text{0.77} & \text{0.78} \\
& \textbf{10} & \text{0.79} & \text{0.81} & \text{0.80} & \text{0.81} & \text{0.80} & \text{0.79} \\
& \textbf{100} & \text{0.79} & \text{0.76} & \text{0.79} & \text{0.77} & \text{0.77} & \text{0.77} \\
\bottomrule
\end{tabular}

\caption{ $LA$ on ImageNet-100 and CUB datasets for different values of $\lambda_1$ and $\lambda_2$.}
\label{tab:lambda-af}
\end{table}

\begin{figure}
    \centering
    \includegraphics[width=0.9\linewidth]{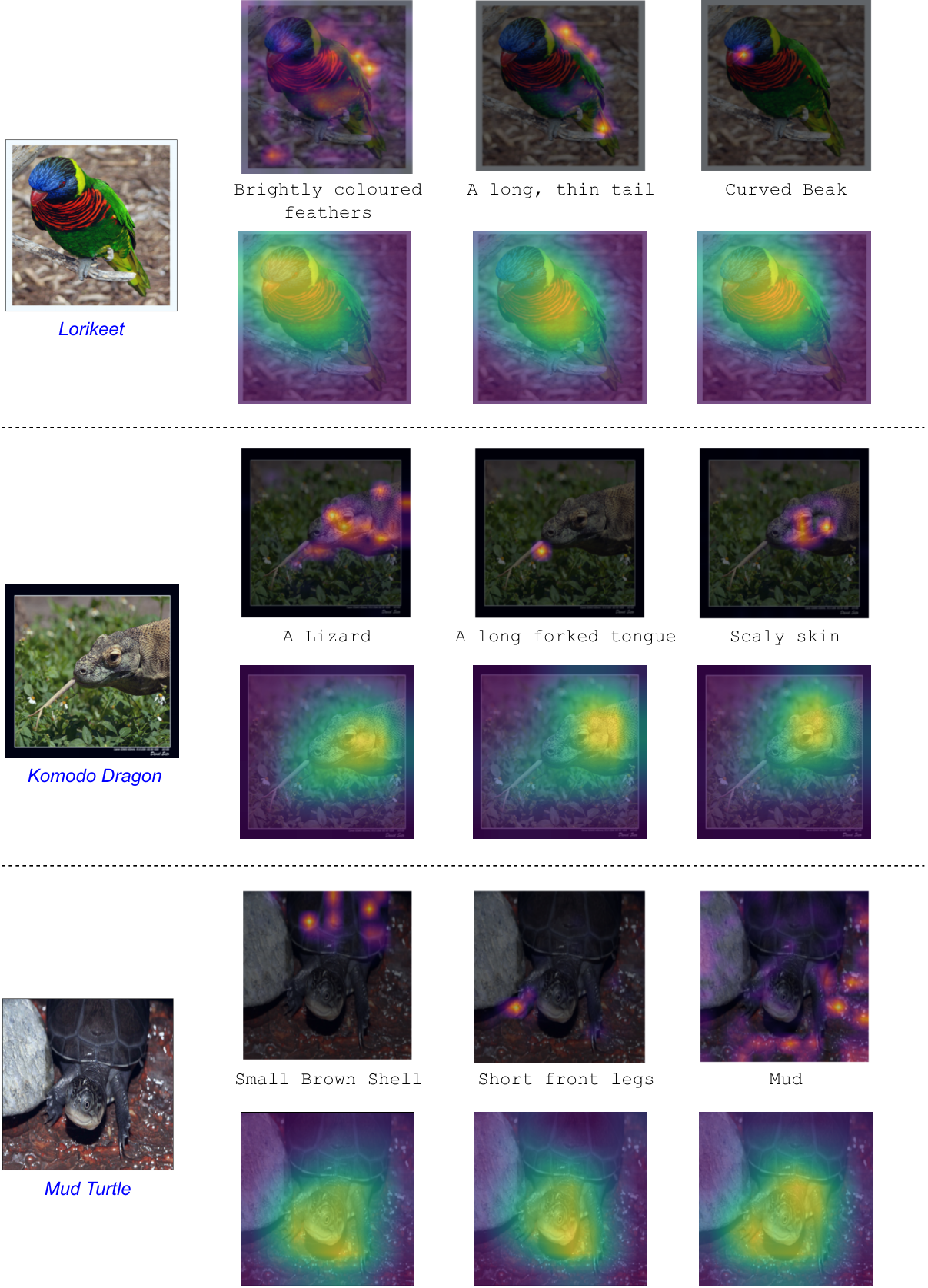}
       \caption{More localization results. Alternate rows present the localization results using \ours versus when localizing the same concepts using GradCAM on CBMs, showing that \ours can provide superior post-hoc visual explanations.}
        \label{fig:appendix-posthoc-localizations}
\end{figure}

\begin{figure}
    \centering
    \includegraphics[width=0.9\linewidth]{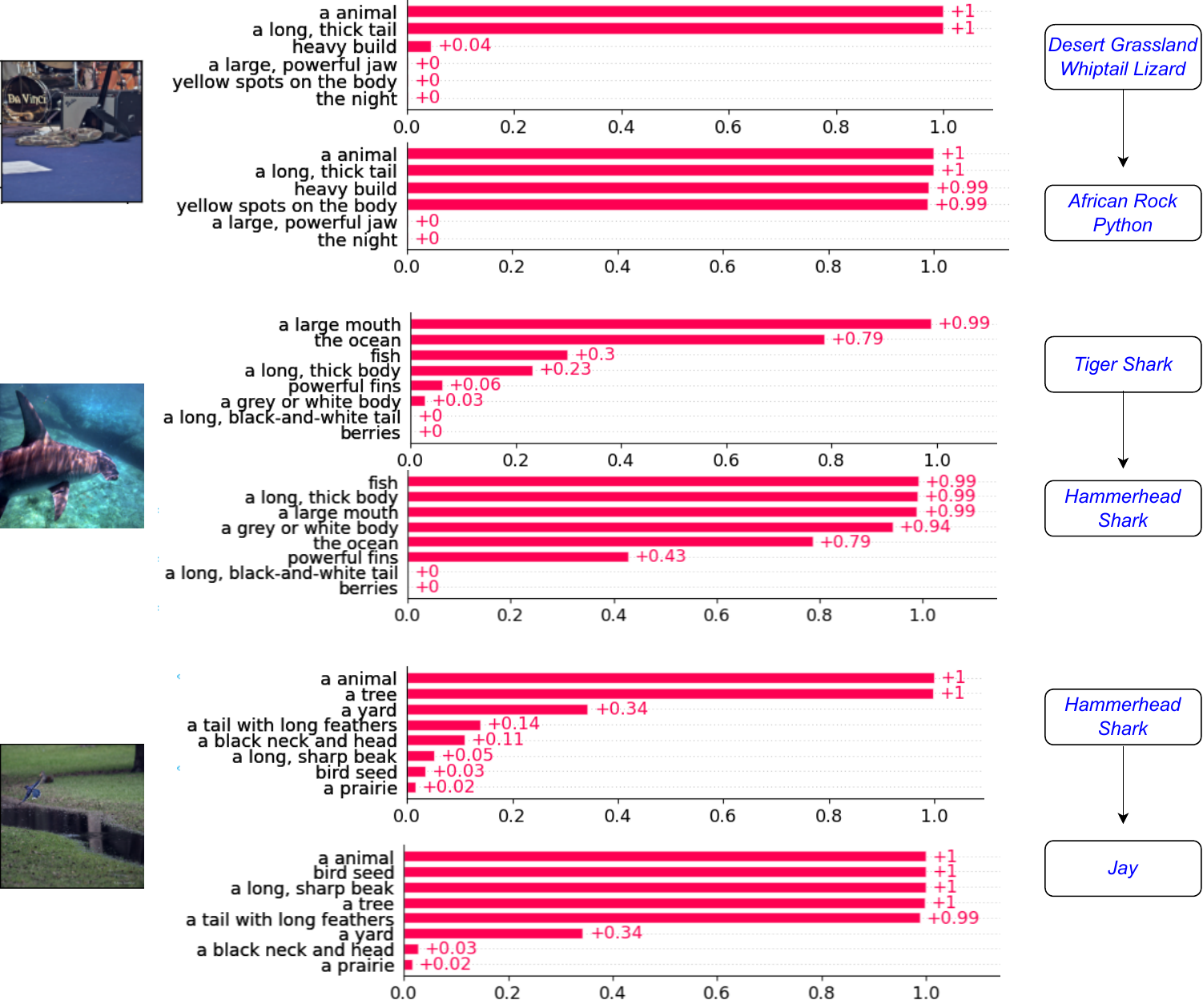}
       \caption{Qualitative results on manual interventions; as illustrated, concept strengths are adjusted to correct class predictions.}
        \label{fig:appendix-posthoc-interventions}
\end{figure}

\noindent \textit{More Qualitative Results.}\label{subsec:more_qualitative_results} More results for attention localization similar to Figure \ref{fig:visual-grounding} are shown in Figure \ref{fig:appendix-posthoc-localizations}. Here 
too, we see that \ours is able to localize specific areas in the input images that relate to certain visual concepts, as opposed to GradCAM on CBMs that produces diffused regions of interest. For interventions, we provide qualitative results in Fig \ref{fig:appendix-posthoc-interventions}. We see that chaning a few key attributes helps the model classify the image correctly. For example, when increasing the strengths of concepts relating to "beak", "tail and feathers", and "tree", we see that the model changes an incorrect prediction from "Hammerhead Shark" to "Jay".

\subsection{Dataset Descriptions} \label{subsec:dataset_details}

\textbf{Cifar100} consists of 50000 training images and 10000 validation images spanning 100 classes. Each image is a 3-channel RGB image of size 32x32 pixels. Concept annotations per class are not provided, and hence we query a Large Language Model as described by \citep{oikarinen2023labelfree} to get the concept set, excluding the concept filters applied post-training for reduction in the number of concepts. We get a 
\begin{table}
    \centering
    \begin{tabular}{@{}lcc@{}}
    \toprule
    \textbf{Exp} & \textbf{CIFAR-100} & \textbf{ImageNet-100} \\
    \midrule
    E1 & 257 (257) & 214 (214) \\
    E2 & 460 (527) & 359 (416) \\
    E3 & 638 (794) & 457 (594) \\
    E4 & 798 (1046) & 545 (762) \\
    E5 & 925 (1309) & 641 (945) \\
    \bottomrule
    \end{tabular}
    \caption{ Number of concepts per class over 5 experiences, excluding duplicates across experiences (Exp) (inclusive numbers in parentheses)}
    \label{tab:concept_count}
\end{table}
total of 925 concepts for Cifar100. \textbf{CUB200 (CalTech-UCSD Birds 200)} is a fine-grained bird identification dataset consisting of 11000 RGB images of 200 different bird species. In this case, the concept annotations are provided by human annotators. 
All concepts are shared among a few classes, which means that the entire concept set is available from the first experience itself. 
This gives us a platform to show that our method gives state-of-the-art results even on fine-grained visual classification in a much simpler setting where existing methods still fail to perform comparably. \textbf{ImageNet100} is a subset of the Imagenet1K dataset consisting of 100 classes with 1300 training images and 50 validation images per class. The subset includes both coarse and fine-grained classes. We choose classes such that each class in a new experience adds new concepts, in addition to using concepts available from past experiences. We provide a subset of some classes and concepts per dataset in \ref{fig:appendix-sampledatasets}. To use the datasets in a continual setting, we split each dataset into 5 or 10 experiences having overlapping concept sets. Details of the number of concepts in each experience for Imagenet100 and Cifar100 for 5 experiences have been provided in Table \ref{tab:concept_count}. In the case of CUB, all experiences share the same set of 312 concepts. Thus, for CUB, we follow the continual learning setting as in \cite{marconato2022catastrophic}.

\begin{table*}[t]
\centering
\begin{tabular}{|c|c|p{6cm}|}
\hline
\textbf{Dataset} & \textbf{Class} & \textbf{Concepts} \\
\hline
\multirow{3}{*}{\parbox{1.5cm}{\vspace{1.15cm} CIFAR100}} & \multirow{1}{*}{\parbox{1cm}{\vspace{0.5cm} Bicycle}} & 
  a tire, object, a helmet, a handlebar, a bicycle seat, pedals attached to the frame, mode of transportation, two wheels of equal size, a seat affixed to the frame, a chain \vspace{3pt} \\
& \multirow{1}{*}{\parbox{0.75cm}{\vspace{0.25cm} Chair}} & 
  furniture, a person, object, legs to support the seat, an office, a computer, a desk, four legs, a backrest, armrests on either side \vspace{3pt} \\
& \multirow{1}{*}{\parbox{1.25cm}{\vspace{0.25cm} Kangaroo}} & 
  a grassland, short front legs, an animal, a safari, mammal, a long, powerful tail, brown or gray fur, marsupial, long, powerful hind legs, Australia\\
\hline
\multirow{3}{*}{\parbox{1.85cm}{\vspace{1.85cm} Imagenet100}} & \multirow{1}{*}{\parbox{1.5cm}{\vspace{0.45cm} Chickadee}}& 
  trees, grayish upperparts, vertebrate, a short beak, white cheeks, chordate, gray wings and back, an animal, leaves, a small, round shape \vspace{3pt} \\
& \multirow{1}{*}{\parbox{2.65cm}{\vspace{0.6cm} American Bullfrog}} & 
  an animal, a stream, a large size, a log, a webbed foot, a marsh, a lily pad, long, powerful hind legs, a large body, a swamp, a carnivorous diet, a lake, a woods, large, webbed hind feet, a large mouth, a river, a pond, spots or blotches on the skin, a green or brown body \vspace{3pt} \\
& \multirow{1}{*}{\parbox{2.45cm}{\vspace{0.5cm} Komodo Dragon}} & 
  a large size, a keeper, scales, a tree, a dish, scaly skin, a rock, long, sharp claws, a long, thick tail, a long, forked tongue, an animal, reptile, a fence, vertebrate, a water dish, a zoo, a heat lamp, a large, bulky body, a cage, a lizard \\
\hline
\multirow{3}{*}{\parbox{0.85cm}{\vspace{1.75cm} \hspace{4cm} CUB}} & \multirow{1}{*}{\parbox{3.25cm}{\vspace{0.5cm} Black-footed Albatross}} & 
  back pattern: solid, under tail color: rufous, wing shape: long-wings, belly color: red, wing color: red, upperparts color: brown, breast pattern: multi-colored, upperparts color: rufous, bill shape: cone, tail shape: notched tail, back color: blue \vspace{3pt} \\
& \multirow{1}{*}{\parbox{2.25cm}{\vspace{0.6cm}  American Crow}} & 
  back pattern: solid, wing shape: long-wings, upperparts color: brown, bill shape: cone, tail shape: notched tail, back color: blue, under tail color: grey, wing shape: tapered-wings, belly color: iridescent, wing color: iridescent \vspace{3pt} \\
& \multirow{1}{*}{\parbox{2.25cm}{\vspace{0.5cm} Lazuli Bunting}} & 
  back pattern: solid, under tail color: rufous, throat color: pink, wing shape: long-wings, wing color: red, upper tail color: pink, upperparts color: brown, breast pattern: multi-colored, bill shape: cone, tail shape: notched tail \\
\hline
\end{tabular}
\caption{ Sample classes and a subset of their corresponding concepts for the three datasets}
\label{fig:appendix-sampledatasets}
\end{table*}

\subsection{Limitations, Future Directions and Broader Impact}\label{subsec:limitations_broader_impact}
The primary dependency of our framework is the need for an LLM to provide the concepts at a class level. We however believe that this has become very feasible in recent times, especially since we only need this at a class level. 
As shown in Section \ref{sec_expts} of the main paper, replay currently plays a very important role in preserving concept-class relationships. It would be interesting to see if exemplar-free continual learning approaches (without replay) can still preserve these relationships, even though they don't have access to labeled concept-class pairs from past experiences. From an interpretability viewpoint, developing an improved intervention mechanism that can be used on our model - even in between experiences, without an explicit linear layer would be an interesting direction of future work.

In this work, we introduced a unique mechanism that enables scaling up the number of concepts and classes without increasing the number of parameters. This method can thus be easily adapted to learn new prototypes continually as well. The multimodal concept embeddings provided by our method can also be extended beyond continual learning applications to incorporate other experiences such as concept-based novel class discovery and concept-based open-world classification. Furthermore, due to the flexibility of our multimodal concept encoder, we can extend our work to process other input modalities such as audio. In such an application, a natural choice of concept anchors could be phoneme representations. Lastly, to the best of our knowledge, our framework does not pose explicit ethical concerns.

\subsection{Per-Experience Performance Results}\label{subsec:per_experience_performance} 
In continuation to the results in Section 4 of the main paper, we herein report the performance of \oursbf and baseline methods across 5 and 10 experiences using ER, providing the values of Average Accuracies and Forgetting at every experience. These numbers are reported for different exemplar buffer sizes: 500, 2000, and 5000. The best-performing method is highlighted across each table.

\begin{table*}[h]
  \centering
  \begin{tabular}{@{}l *{10}{c}@{}}
    \toprule
    \textbf{Model} & \multicolumn{2}{c}{\textbf{Exp 1}} & \multicolumn{2}{c}{\textbf{Exp 2}} & \multicolumn{2}{c}{\textbf{Exp 3}} & \multicolumn{2}{c}{\textbf{Exp 4}} & \multicolumn{2}{c}{\textbf{Exp 5}} \\
    \cmidrule(lr){2-3} \cmidrule(lr){4-5} \cmidrule(lr){6-7} \cmidrule(lr){8-9} \cmidrule(l){10-11}
    & \textbf{FAA} & \textbf{AF} & \textbf{FAA} & \textbf{AF} & \textbf{FAA} & \textbf{AF} & \textbf{FAA} & \textbf{AF} & \textbf{FAA} & \textbf{AF} \\
    \midrule
    CBM-S & 0.9177 & N/A & 0.6997 & 0.7691 & 0.4949 & 0.7669 & 0.3956 & 0.6652 & 0.3005 & 0.5459  \\
    CBM-J & 0.8497 & N/A & 0.5641 & 0.8105 & 0.3876 & 0.7696 & 0.3004 & 0.7269 & 0.2293 & 0.6513 \\
    ICIAP-S & 0.8745 & N/A & 0.5598 & 0.7842 & 0.3831 & 0.6763 & 0.2738 & 0.5945 & 0.2132 & 0.4345 \\
    ICIAP-J & 0.8449 & N/A & 0.5719 & 0.7785 & 0.3971 & 0.7691 & 0.2981 & 0.7192 & 0.2450 & 0.6304 \\
    LabelFree & 0.1847 & N/A & 0.2033 & 0.6675 & 0.1985 & 0.3915 & 0.2030 & \textbf{0.2443} & 0.2172 & \textbf{0.0675} \\
    LaBo & 0.9065 & N/A & 0.6999 & 0.9401 & 0.5679 & 0.7381 & 0.4558 & 0.7648 & 0.3004 & 0.6016 \\
    \midrule
    \oursbf & \textbf{0.9545} & N/A &  \textbf{0.8069} & \textbf{0.3795} & \textbf{0.7391} & \textbf{0.3545} & \textbf{0.7078} & 0.3795 & \textbf{0.6657} & 0.2789 \\
    \bottomrule
  \end{tabular}
  \caption{Per Experience Results for CIFAR100 with 500 Exemplars and 5 Experiences}
  \label{CIFAR100_500Exemplar}
\end{table*}

\begin{table*}[h]
  \centering
  \begin{tabular}{@{}l *{10}{c}@{}}
    \toprule
    \textbf{Model} & \multicolumn{2}{c}{\textbf{Exp 1}} & \multicolumn{2}{c}{\textbf{Exp 2}} & \multicolumn{2}{c}{\textbf{Exp 3}} & \multicolumn{2}{c}{\textbf{Exp 4}} & \multicolumn{2}{c}{\textbf{Exp 5}} \\
    \cmidrule(lr){2-3} \cmidrule(lr){4-5} \cmidrule(lr){6-7} \cmidrule(lr){8-9} \cmidrule(l){10-11}
    & \textbf{FAA} & \textbf{AF} & \textbf{FAA} & \textbf{AF} & \textbf{FAA} & \textbf{AF} & \textbf{FAA} & \textbf{AF} & \textbf{FAA} & \textbf{AF} \\
    \midrule
    CBM-S & 0.9177 & N/A & 0.7971 & 0.5826 & 0.6590 & 0.4797 & 0.5660 & 0.4385 & 0.4698 & 0.3782 \\
    CBM-J & 0.8497 & N/A & 0.6761 & 0.5985 & 0.5351 & 0.5682 & 0.4581 & 0.5508 & 0.3855 & 0.4638 \\
    ICIAP-S & 0.8688 & N/A & 0.6879 & 0.5690 & 0.5639 & 0.4120 & 0.4632 & 0.3977 & 0.3937 & 0.3157 \\
    ICIAP-J & 0.8229 & N/A & 0.6762 & 0.6082 & 0.5468 & 0.5687 & 0.4742 & 0.4938 & 0.3854 & 0.4391 \\
    LabelFree & 0.3013 & N/A & 0.3121 & 0.5504 & 0.3133 &
    \textbf{0.2628} & 0.3238 & \textbf{0.1246} & 0.3398 &
    \textbf{0.0314} \\
    LaBo & 0.9076 & N/A & 0.7123 & 0.7908 & 0.5888 & 0.6485 & 0.4612 & 0.7004 & 0.3048 & 0.5600 \\
    \midrule
    \oursbf & \textbf{0.9559} & N/A & \textbf{0.8470} & \textbf{0.3097} & \textbf{0.7860} & 0.3017 & \textbf{0.7452} & 0.3011 & \textbf{0.7120} & 0.2372 \\
    \bottomrule
  \end{tabular}
  \caption{Per Experience Results for CIFAR100 with 2000 Exemplars and 5 Experiences}
  \label{CIFAR100_2000Exemplar}
\end{table*}

\begin{table*}[h]
  \centering
  \begin{tabular}{@{}l *{10}{c}@{}}
    \toprule
    \textbf{Model} & \multicolumn{2}{c}{\textbf{Exp 1}} & \multicolumn{2}{c}{\textbf{Exp 2}} & \multicolumn{2}{c}{\textbf{Exp 3}} & \multicolumn{2}{c}{\textbf{Exp 4}} & \multicolumn{2}{c}{\textbf{Exp 5}} \\
    \cmidrule(lr){2-3} \cmidrule(lr){4-5} \cmidrule(lr){6-7} \cmidrule(lr){8-9} \cmidrule(l){10-11}
    & \textbf{FAA} & \textbf{AF} & \textbf{FAA} & \textbf{AF} & \textbf{FAA} & \textbf{AF} & \textbf{FAA} & \textbf{AF} & \textbf{FAA} & \textbf{AF} \\
    \midrule
    CBM-S & 0.9177 & N/A & 0.8471 & 0.4077 & 0.7245 & 0.3661 & 0.6307 & 0.3142 & 0.5724 & 0.2279 \\
    CBM-J & 0.8497 & N/A & 0.7398 & 0.4353 & 0.6217 & 0.4080 & 0.5683 & 0.3507 & 0.4969 & 0.3383 \\
    ICIAP-S & 0.8745 & N/A & 0.7489 & 0.4348 & 0.6247 & 0.3259 & 0.5373 & 0.2537 & 0.4587 & 0.2688 \\
    ICIAP-J & 0.8450 & N/A & 0.7323 & 0.3976 & 0.6384 & 0.3547 & 0.5571 & 0.3425 & 0.5160 & 0.3544 \\
    LabelFree & 0.3815 & N/A & 0.3713 & 0.4710 & 0.3905 &
    \textbf{0.2472} & 0.3975 & \textbf{0.0988} & 0.4027 &
    \textbf{0.0330} \\
    LaBo & 0.9085 & N/A & 0.7151 & 0.7663 & 0.5969 & 0.6315 & 0.4674 & 0.6927 & 0.3095 & 0.5258 \\
    \midrule
    \oursbf & \textbf{0.9560} & N/A & \textbf{0.8849} & \textbf{0.2614} & \textbf{0.8177} & \textbf{0.2665} & \textbf{0.7805} & 0.2644 & \textbf{0.7425} & 0.1963 \\
    \bottomrule
  \end{tabular}
  \caption{Per Experience Results for CIFAR100 with 5000 Exemplars and 5 Experiences}
  \label{CIFAR100_5000Exemplar}
\end{table*}

\begin{table*}[h]
  \centering
  \begin{tabular}{@{}l *{10}{c}@{}}
    \toprule
    \textbf{Model} & \multicolumn{2}{c}{\textbf{Exp 1}} & \multicolumn{2}{c}{\textbf{Exp 2}} & \multicolumn{2}{c}{\textbf{Exp 3}} & \multicolumn{2}{c}{\textbf{Exp 4}} & \multicolumn{2}{c}{\textbf{Exp 5}} \\
    \cmidrule(lr){2-3} \cmidrule(lr){4-5} \cmidrule(lr){6-7} \cmidrule(lr){8-9} \cmidrule(l){10-11}
    & \textbf{FAA} & \textbf{AF} & \textbf{FAA} & \textbf{AF} & \textbf{FAA} & \textbf{AF} & \textbf{FAA} & \textbf{AF} & \textbf{FAA} & \textbf{AF} \\
    \midrule
    CBM-Seq & 0.7121 & N/A & 0.5967 & 0.5383 & 0.5033 & 0.4463 & 0.4038 & 0.4517 & 0.3494 & 0.3854 \\
    CBM-J & 0.6845 & N/A & 0.5569 & 0.4656 & 0.4765 & 0.3463 & 0.4068 & 0.3358 & 0.3753 & 0.2992 \\
    ICIAP-S & 0.7121 & N/A & 0.5967 & 0.5383 & 0.5033 & 0.4463 & 0.4038 & 0.4517 & 0.3494 & 0.3854 \\
    ICIAP-J & 0.6845 & N/A & 0.5569 & 0.4656 & 0.4765 & 0.3463 & 0.4068 & 0.3358 & 0.3753 & 0.2992 \\
    LabelFree & 0.0557 & N/A & 0.1049 & 0.6942 & 0.0751 & 0.2910 & 0.0529 & 0.2858 & 0.3121 & 0.2497 \\
    LaBo & 0.6435 & N/A & 0.5869 & 0.5532 & 0.5606 & 0.5030 & 0.3942 & 0.6065 & 0.2869 & 0.6107 \\
    \midrule
    \oursbf & \textbf{0.8655} & N/A & \textbf{0.8246} & \textbf{0.1539} & \textbf{0.8245} & \textbf{0.0722} & \textbf{0.7782} & \textbf{0.1086} & \textbf{0.7832} & \textbf{0.1151} \\
    \bottomrule
  \end{tabular}
  \caption{Per Experience Results for Per Experience Results for CUB (Caltech-UCSD
Birds-200-2011) with 500 Exemplars and 5 Experiences}
  \label{CUB_500Exemplar}
\end{table*}

\begin{table*}[h]
  \centering
  \begin{tabular}{@{}l *{10}{c}@{}}
    \toprule
    \textbf{Model} & \multicolumn{2}{c}{\textbf{Exp 1}} & \multicolumn{2}{c}{\textbf{Exp 2}} & \multicolumn{2}{c}{\textbf{Exp 3}} & \multicolumn{2}{c}{\textbf{Exp 4}} & \multicolumn{2}{c}{\textbf{Exp 5}} \\
    \cmidrule(lr){2-3} \cmidrule(lr){4-5} \cmidrule(lr){6-7} \cmidrule(lr){8-9} \cmidrule(l){10-11}
    & \textbf{FAA} & \textbf{AF} & \textbf{FAA} & \textbf{AF} & \textbf{FAA} & \textbf{AF} & \textbf{FAA} & \textbf{AF} & \textbf{FAA} & \textbf{AF} \\
    \midrule
    CBM-Seq & 0.7121 & N/A & 0.6667 & 0.2654 & 0.6041 & 0.1675 & 0.5450 & 0.2033 & 0.5332 & 0.1796 \\
    CBM-J & 0.6845 & N/A & 0.6218 & 0.1452 & 0.5759 & 0.0850 & 0.5473 & 0.1008 & 0.5566 & 0.0917 \\
    ICIAP-S & 0.7121 & N/A & 0.6667 & 0.2654 & 0.6041 & 0.1675 & 0.5450 & 0.2033 & 0.5332 & 0.1796 \\
    ICIAP-J & 0.6845 & N/A & 0.6218 & 0.1452 & 0.5759 & 0.0850 & 0.5473 & 0.1008 & 0.5566 & 0.0917 \\
    LabelFree & 0.1807 & N/A & 0.2104 & 0.5700 & 0.0875 & 0.3257 & 0.0000 & 0.4436 & 0.4508 & 0.4434 \\
    LaBo & 0.6449 & N/A & 0.6021 & 0.4912 & 0.574 & 0.4365 & 0.4204 & 0.5336 & 0.3062 & 0.5205 \\
    \midrule
    \oursbf & \textbf{0.8591} & N/A & \textbf{0.8356} & \textbf{0.0787} & \textbf{0.8458} & \textbf{0.0297} & \textbf{0.8195} & \textbf{0.0680} & \textbf{0.8215} & \textbf{0.0551} \\
    \bottomrule
  \end{tabular}
  \caption{Per Experience Results for Per Experience Results for CUB (Caltech-UCSD
Birds-200-2011) with 2000 Exemplars and 5 Experiences}
  \label{CUB_2000Exemplar}
\end{table*}

\begin{table*}[h]
  \centering
  \begin{tabular}{@{}l *{10}{c}@{}}
    \toprule
    \textbf{Model} & \multicolumn{2}{c}{\textbf{Experience 1}} & \multicolumn{2}{c}{\textbf{Experience 2}} & \multicolumn{2}{c}{\textbf{Experience 3}} & \multicolumn{2}{c}{\textbf{Experience 4}} & \multicolumn{2}{c}{\textbf{Experience 5}} \\
    \cmidrule(lr){2-3} \cmidrule(lr){4-5} \cmidrule(lr){6-7} \cmidrule(lr){8-9} \cmidrule(l){10-11}
    & \textbf{FAA} & \textbf{AF} & \textbf{FAA} & \textbf{AF} & \textbf{FAA} & \textbf{AF} & \textbf{FAA} & \textbf{AF} & \textbf{FAA} & \textbf{AF} \\
    \midrule
    CBM-Seq & 0.7121 & N/A & 0.6667 & 0.1806 & 0.6147 & 0.0937 & 0.5792 & 0.0808 & 0.5915 & 0.0721 \\
    CBM-J & 0.6845 & N/A & 0.6218 & 0.1178 & 0.5832 & 0.0255 & 0.5558 & 0.0500 & 0.5716 & 0.0575 \\
    ICIAP-S & 0.7121 & N/A & 0.6667 & 0.1806 & 0.6147 & 0.0937 & 0.5792 & 0.0808 & 0.5915 & 0.0721 \\
    ICIAP-J & 0.6845 & N/A & 0.6218 & 0.1178 & 0.5832 & 0.0255 & 0.5558 & 0.0500 & 0.5716 & 0.0575 \\
    LabelFree & 0.4211 & N/A & 0.0000 & 0.3211 & 0.0000 & 0.5338 & 0.0000 & 0.5276 & 0.4664 & 0.4569 \\
    LaBo & 0.6460 & N/A & 0.6322 & 0.4594 & 0.5382 & 0.3897 & 0.4472 & 0.4897 & 0.3121 & 0.4896 \\
    \midrule
    \oursbf & \textbf{0.8432} & N/A & \textbf{0.8414} & \textbf{0.0472} & \textbf{0.8446} & \textbf{0.0158} & \textbf{0.8229} & \textbf{0.0322} & \textbf{0.8333} & \textbf{0.0274} \\
    \bottomrule
  \end{tabular}
  \caption{Per Experience Results for Per Experience Results for CUB (Caltech-UCSD
Birds-200-2011) with 5000 Exemplars and 5 Experiences}
  \label{CUB_5000Exemplar}
\end{table*}

\begin{table*}[h]
  \centering
  \begin{tabular}{@{}l *{10}{c}@{}}
    \toprule
    \textbf{Model} & \multicolumn{2}{c}{\textbf{Experience 1}} & \multicolumn{2}{c}{\textbf{Experience 2}} & \multicolumn{2}{c}{\textbf{Experience 3}} & \multicolumn{2}{c}{\textbf{Experience 4}} & \multicolumn{2}{c}{\textbf{Experience 5}} \\
    \cmidrule(lr){2-3} \cmidrule(lr){4-5} \cmidrule(lr){6-7} \cmidrule(lr){8-9} \cmidrule(l){10-11}
    & \textbf{FAA} & \textbf{AF} & \textbf{FAA} & \textbf{AF} & \textbf{FAA} & \textbf{AF} & \textbf{FAA} & \textbf{AF} & \textbf{FAA} & \textbf{AF} \\
    \midrule
    CBM-S & 0.9246 & N/A & 0.6547 & 0.7484 & 0.5028 & 0.6569 & 0.3380 & 0.5867 & 0.2904 & 0.4655 \\
    CBM-J & 0.8605 & N/A & 0.5548 & 0.6657 & 0.4313 & 0.5960 & 0.3277 & 0.5865 & 0.3040 & 0.4990 \\
    ICIAP-S & 0.8827 & N/A & 0.5855 & 0.7702 & 0.4136 & 0.5246 & 0.2968 & 0.4704 & 0.2199 & 0.4399 \\
    ICIAP-J & 0.8536 & N/A & 0.5781 & 0.7099 & 0.4089 & 0.5413 & 0.2968 & 0.5615 & 0.3100 & 0.4788 \\
    LabelFree & 0.0813 & N/A & 0.0593 & 0.8393 & 0.1107 & 0.3600 & 0.0667 & \textbf{0.0360} & 0.0716 &        \textbf{0.0000} \\
    LaBo & 0.5721 & N/A & 0.3547 & 0.5466 & 0.4116 & 0.4543 & 0.3891 & 0.4543 & 0.4071 & 0.6453 \\
    \midrule
    \oursbf & \textbf{0.9403} & N/A & \textbf{0.8563} & \textbf{0.1017} & \textbf{0.8119} & \textbf{0.1098} & \textbf{0.7368} & 0.1193 & \textbf{0.7975} & 0.0261 \\
    \bottomrule
  \end{tabular}
  \caption{Per Experience Results for ImageNet100 with 500 Exemplars and 5 Experiences}
  \label{ImageNet100_500Exemplar}
\end{table*}

\begin{table*}[h]
  \centering
  \begin{tabular}{@{}l *{10}{c}@{}}
    \toprule
    \textbf{Model} & \multicolumn{2}{c}{\textbf{Experience 1}} & \multicolumn{2}{c}{\textbf{Experience 2}} & \multicolumn{2}{c}{\textbf{Experience 3}} & \multicolumn{2}{c}{\textbf{Experience 4}} & \multicolumn{2}{c}{\textbf{Experience 5}} \\
    \cmidrule(lr){2-3} \cmidrule(lr){4-5} \cmidrule(lr){6-7} \cmidrule(lr){8-9} \cmidrule(l){10-11}
    & \textbf{FAA} & \textbf{AF} & \textbf{FAA} & \textbf{AF} & \textbf{FAA} & \textbf{AF} & \textbf{FAA} & \textbf{AF} & \textbf{FAA} & \textbf{AF} \\
    \midrule
    CBM-S & 0.9246 & N/A & 0.7753 & 0.4383 & 0.6216 & 0.3980 & 0.5255 & 0.2962 & 0.5079 & 0.2490 \\
    CBM-J & 0.8605 & N/A & 0.6417 & 0.4446 & 0.5584 & 0.3825 & 0.4373 & 0.4399 & 0.4588 & 0.3169 \\
    ICIAP-S & 0.8827 & N/A & 0.6732 & 0.4611 & 0.5536 & 0.3379 & 0.4364 & 0.3109 & 0.4095 & 0.2071 \\
    ICIAP-J & 0.8536 & N/A & 0.6751 & 0.4415 & 0.5691 & 0.3875 & 0.4385 & 0.4308 & 0.4676 & 0.3151 \\
    LabelFree & 0.1500 & N/A & 0.1133 & 0.7700 & 0.1307 & 0.2083 & 0.1333 & 0.2270 & 0.0944 & \textbf{0.0010} \\
    LaBo & 0.5746 & N/A & 0.3757 & 0.5310 & 0.4402 & 0.3963 & 0.429 & 0.3744 & 0.4353 & 0.6064 \\
    \midrule
    \oursbf & \textbf{0.9347} & N/A & \textbf{0.8606} & \textbf{0.1116} & \textbf{0.8106} & \textbf{0.1141} & \textbf{0.7526} & \textbf{0.1362} & \textbf{0.7924} & 0.0161 \\
    \bottomrule
  \end{tabular}
  \caption{Per Experience Results for ImageNet100 with 2000 Exemplars and 5 Experiences}
  \label{ImageNet100_2000Exemplar}
\end{table*}

\begin{table*}[h]
  \centering
  \begin{tabular}{@{}l *{10}{c}@{}}
    \toprule
    \textbf{Model} & \multicolumn{2}{c}{\textbf{Experience 1}} & \multicolumn{2}{c}{\textbf{Experience 2}} & \multicolumn{2}{c}{\textbf{Experience 3}} & \multicolumn{2}{c}{\textbf{Experience 4}} & \multicolumn{2}{c}{\textbf{Experience 5}} \\
    \cmidrule(lr){2-3} \cmidrule(lr){4-5} \cmidrule(lr){6-7} \cmidrule(lr){8-9} \cmidrule(l){10-11}
    & \textbf{FAA} & \textbf{AF} & \textbf{FAA} & \textbf{AF} & \textbf{FAA} & \textbf{AF} & \textbf{FAA} & \textbf{AF} & \textbf{FAA} & \textbf{AF} \\
    \midrule
    CBM-S & 0.9246 & N/A & 0.8032 & 0.3329 & 0.6950 & 0.2766 & 0.6029 & 0.2077 & 0.5936 & 0.1516 \\
    CBM-J & 0.8605 & N/A & 0.7212 & 0.3337 & 0.6210 & 0.3331 & 0.5222 & 0.3427 & 0.5289 & 0.2175 \\
    ICIAP-S & 0.8827 & N/A & 0.7213 & 0.3024 & 0.6039 & 0.2423 & 0.5346 & 0.1498 & 0.5193 & 0.2147 \\
    ICIAP-J & 0.8536 & N/A & 0.7074 & 0.2933 & 0.6225 & 0.3129 & 0.5168 & 0.3256 & 0.5465 & 0.2127 \\
    LabelFree & 0.2987 & N/A & 0.2633 & 0.6233 & 0.3213 & \textbf{0.0507} & 0.2853 & \textbf{0.1000} & 0.1782 & \textbf{0.0000} \\
    LaBo & 0.5736 & N/A & 0.3723 & 0.5037 & 0.4344 & 0.3746 & 0.4362 & 0.3574 & 0.4305 & 0.5926 \\
    \midrule
    \oursbf & \textbf{0.9354} & N/A & \textbf{0.8652} & \textbf{0.0831} & \textbf{0.8153} & 0.1036 & \textbf{0.7626} & 0.1046 & \textbf{0.8057} & 0.0265 \\
    \bottomrule
  \end{tabular}
  \caption{Per Experience Results for ImageNet100 with 5000 Exemplars and 5 Experiences}
  \label{ImageNet100_5000Exemplar}
\end{table*}

\begin{table*}[h]
  \centering
  \setlength\tabcolsep{2pt} 
  \begin{tabular}{@{}l *{20}{c}@{}}
    \toprule
    \textbf{Model} & \multicolumn{2}{c}{\textbf{Exp 1}} & \multicolumn{2}{c}{\textbf{Exp 2}} & \multicolumn{2}{c}{\textbf{Exp 3}} & \multicolumn{2}{c}{\textbf{Exp 4}} & \multicolumn{2}{c}{\textbf{Exp 5}} & \multicolumn{2}{c}{\textbf{Exp 6}} & \multicolumn{2}{c}{\textbf{Exp 7}} & \multicolumn{2}{c}{\textbf{Exp 8}} & \multicolumn{2}{c}{\textbf{Exp 9}} & \multicolumn{2}{c}{\textbf{Exp 10}} \\
    \cmidrule(lr){2-3} \cmidrule(lr){4-5} \cmidrule(lr){6-7} \cmidrule(lr){8-9} \cmidrule(lr){10-11} \cmidrule(lr){12-13} \cmidrule(lr){14-15} \cmidrule(lr){16-17} \cmidrule(lr){18-19} \cmidrule(l){20-21}
    & \textbf{FAA} & \textbf{AF} & \textbf{FAA} & \textbf{AF} & \textbf{FAA} & \textbf{AF} & \textbf{FAA} & \textbf{AF} & \textbf{FAA} & \textbf{AF} & \textbf{FAA} & \textbf{AF} & \textbf{FAA} & \textbf{AF} & \textbf{FAA} & \textbf{AF} & \textbf{FAA} & \textbf{AF} & \textbf{FAA} & \textbf{AF} \\
    \midrule
    CBM-S & 0.95 & N/A & 0.80 & 0.75 & 0.71 & 0.74 & 0.63 & 0.69 & 0.50 & 0.83 & 0.46 & 0.72 & 0.41 & 0.71 & 0.36 & 0.64 & 0.33 & 0.70 & 0.28 & 0.66 \\
    CBM-J & 0.91 & N/A & 0.65 & 0.85 & 0.50 & 0.85 & 0.43 & 0.80 & 0.32 & 0.86 & 0.29 & 0.82 & 0.26 & 0.75 & 0.23 & 0.80 & 0.20 & 0.77 & 0.15 & 0.77 \\
    ICIAP-S & 0.92 & N/A & 0.69 & 0.76 & 0.60 & 0.77 & 0.51 & 0.70 & 0.39 & 0.79 & 0.34 & 0.68 & 0.30 & 0.63 & 0.26 & 0.60 & 0.22 & 0.66 & 0.20 & 0.65 \\
    ICIAP-J & 0.91 & N/A & 0.64 & 0.85 & 0.50 & 0.84 & 0.40 & 0.82 & 0.32 & 0.85 & 0.28 & 0.82 & 0.24 & 0.85 & 0.19 & 0.80 & 0.19 & 0.82 & 0.13 & 0.75 \\
    LabelFree & 0.16 & N/A & 0.18 & 0.76 & 0.15 & \textbf{0.39} & 0.18 & 0.34 & 0.16 & \textbf{0.18} & 0.16 & \textbf{0.21} & 0.19 & \textbf{0.11} & 0.16 & \textbf{0.05} & 0.17 & \textbf{0.07} & 0.19 &
    \textbf{0.06} \\
    LaBo & 0.94 & N/A & 0.18 & 0.94 & 0.38 & 0.53 & 0.32 & 0.65 & 0.25 & 0.74 & 0.13 & 0.86 & 0.13 & 0.83 & 0.12 & 0.81 & 0.11 & 0.86 & 0.10 & 0.97 \\
    \midrule
    \oursbf & \textbf{0.98} & N/A & \textbf{0.88} & \textbf{0.39} & \textbf{0.81} & 0.44 & \textbf{0.77} & \textbf{0.35} & \textbf{0.72} & 0.39 & \textbf{0.71} & 0.46 & \textbf{0.70} & 0.35 & \textbf{0.65} & 0.37 & \textbf{0.70} & 0.41 & \textbf{0.63} & 0.25 \\
    \bottomrule
  \end{tabular}
  \caption{Per Experience Results for CIFAR100 with 500 Exemplars and 10 Experiences}
  \label{your-label}
\end{table*}

\begin{table*}[!ht]
  \centering
  \setlength\tabcolsep{2pt} 
  \begin{tabular}{@{}l *{20}{c}@{}}
    \toprule
    \textbf{Model} & \multicolumn{2}{c}{\textbf{Exp 1}} & \multicolumn{2}{c}{\textbf{Exp 2}} & \multicolumn{2}{c}{\textbf{Exp 3}} & \multicolumn{2}{c}{\textbf{Exp 4}} & \multicolumn{2}{c}{\textbf{Exp 5}} & \multicolumn{2}{c}{\textbf{Exp 6}} & \multicolumn{2}{c}{\textbf{Exp 7}} & \multicolumn{2}{c}{\textbf{Exp 8}} & \multicolumn{2}{c}{\textbf{Exp 9}} & \multicolumn{2}{c}{\textbf{Exp 10}} \\
    \cmidrule(lr){2-3} \cmidrule(lr){4-5} \cmidrule(lr){6-7} \cmidrule(lr){8-9} \cmidrule(lr){10-11} \cmidrule(lr){12-13} \cmidrule(lr){14-15} \cmidrule(lr){16-17} \cmidrule(lr){18-19} \cmidrule(l){20-21}
    & \textbf{FAA} & \textbf{AF} & \textbf{FAA} & \textbf{AF} & \textbf{FAA} & \textbf{AF} & \textbf{FAA} & \textbf{AF} & \textbf{FAA} & \textbf{AF} & \textbf{FAA} & \textbf{AF} & \textbf{FAA} & \textbf{AF} & \textbf{FAA} & \textbf{AF} & \textbf{FAA} & \textbf{AF} & \textbf{FAA} & \textbf{AF} \\
    \midrule
    CBM-S & 0.95 & N/A & 0.89 & 0.52 & 0.81 & 0.53 & 0.76 & 0.42 & 0.70 & 0.60 & 0.65 & 0.50 & 0.60 & 0.41 & 0.55 & 0.46 & 0.54 & 0.49 & 0.49 & 0.32 \\
    CBM-J & 0.91 & N/A & 0.76 & 0.62 & 0.68 & 0.58 & 0.60 & 0.55 & 0.53 & 0.60 & 0.48 & 0.65 & 0.44 & 0.56 & 0.41 & 0.57 & 0.39 & 0.61 & 0.34 & 0.53 \\
    ICIAP-S & 0.92 & N/A & 0.75 & 0.65 & 0.72 & 0.44 & 0.66 & 0.44 & 0.61 & 0.55 & 0.55 & 0.53 & 0.52 & 0.48 & 0.44 & 0.46 & 0.43 & 0.46 & 0.37 & 0.42 \\
    ICIAP-J & 0.91 & N/A & 0.78 & 0.64 & 0.68 & 0.60 & 0.60 & 0.62 & 0.52 & 0.65 & 0.48 & 0.60 & 0.44 & 0.62 & 0.39 & 0.61 & 0.40 & 0.60 & 0.32 & 0.59 \\
    LabelFree & 0.29 & N/A & 0.29 & 0.63 & 0.26 &   \textbf{0.30} & 0.30 & \textbf{0.29} & 0.29 &    \textbf{0.18} & 0.28 & \textbf{0.15} & 0.29 &     \textbf{0.08} & 0.27 & \textbf{0.05} & 0.29 &    \textbf{0.04} & 0.29 & \textbf{0.01} \\
    LaBo & 0.93 & N/A & 0.20 & 0.93 & 0.38 & 0.47 & 0.33 & 0.63 & 0.26 & 0.74 & 0.14 & 0.84 & 0.13 & 0.83 & 0.12 & 0.85 & 0.10 & 0.89 & 0.10 & 0.94 \\
    \midrule
    \oursbf & \textbf{0.98} & N/A & \textbf{0.91} & \textbf{0.32} & \textbf{0.85} & \textbf{0.38} & \textbf{0.81} & 0.30 & \textbf{0.77} & 0.34 & \textbf{0.77} & 0.44 & \textbf{0.74} & 0.30 &    \textbf{0.72} & 0.34 & \textbf{0.73} & 0.30 &    \textbf{0.67} & 0.25 \\
    \bottomrule
  \end{tabular}
  \caption{Per Experience Results for CIFAR100 with 2000 Exemplars and 10 Experiences}
  \label{your-label}
\end{table*}

\begin{table*}[!ht]
  \centering
  \setlength\tabcolsep{2pt}  
  \begin{tabular}{@{}l *{20}{c}@{}}
    \toprule
    \textbf{Model} & \multicolumn{2}{c}{\textbf{Exp 1}} & \multicolumn{2}{c}{\textbf{Exp 2}} & \multicolumn{2}{c}{\textbf{Exp 3}} & \multicolumn{2}{c}{\textbf{Exp 4}} & \multicolumn{2}{c}{\textbf{Exp 5}} & \multicolumn{2}{c}{\textbf{Exp 6}} & \multicolumn{2}{c}{\textbf{Exp 7}} & \multicolumn{2}{c}{\textbf{Exp 8}} & \multicolumn{2}{c}{\textbf{Exp 9}} & \multicolumn{2}{c}{\textbf{Exp 10}} \\
    \cmidrule(lr){2-3} \cmidrule(lr){4-5} \cmidrule(lr){6-7} \cmidrule(lr){8-9} \cmidrule(lr){10-11} \cmidrule(lr){12-13} \cmidrule(lr){14-15} \cmidrule(lr){16-17} \cmidrule(lr){18-19} \cmidrule(l){20-21}
    & \textbf{FAA} & \textbf{AF} & \textbf{FAA} & \textbf{AF} & \textbf{FAA} & \textbf{AF} & \textbf{FAA} & \textbf{AF} & \textbf{FAA} & \textbf{AF} & \textbf{FAA} & \textbf{AF} & \textbf{FAA} & \textbf{AF} & \textbf{FAA} & \textbf{AF} & \textbf{FAA} & \textbf{AF} & \textbf{FAA} & \textbf{AF} \\
    \midrule
    CBM-S & 0.95 & N/A & 0.91 & 0.42 & 0.87 & 0.39 & 0.82 & 0.29 & 0.78 & 0.49 & 0.73 & 0.45 & 0.66 & 0.39 & 0.60 & 0.34 & 0.62 & 0.38 & 0.54 & 0.34 \\
    CBM-J & 0.91 & N/A & 0.82 & 0.48 & 0.75 & 0.41 & 0.68 & 0.39 & 0.60 & 0.51 & 0.60 & 0.44 & 0.55 & 0.46 & 0.51 & 0.46 & 0.50 & 0.46 & 0.45 & 0.36 \\
    ICIAP-S & 0.92 & N/A & 0.81 & 0.45 & 0.76 & 0.31 & 0.72 & 0.32 & 0.66 & 0.50 & 0.61 & 0.35 & 0.59 & 0.37 & 0.55 & 0.35 & 0.53 & 0.44 & 0.47 & 0.36 \\
    ICIAP-J & 0.91 & N/A & 0.84 & 0.46 & 0.76 & 0.37 & 0.69 & 0.38 & 0.64 & 0.46 & 0.59 & 0.43 & 0.57 & 0.44 & 0.51 & 0.40 & 0.50 & 0.41 & 0.47 & 0.42 \\
    LabelFree & 0.32 & N/A  & 0.33 & 0.59 & 0.30 &
    \textbf{0.29} & 0.34 & \textbf{0.26} & 0.33 & \textbf{0.17} & 0.33 & \textbf{0.12} & 0.34 & \textbf{0.07} & 0.32 & \textbf{0.07} & 0.32 & \textbf{0.04} & 0.33 & \textbf{0.03} \\
    LaBo & 0.92 & N/A  & 0.24 & 0.92 & 0.40 & 0.41 & 0.36 & 0.60 & 0.29 & 0.72 & 0.18 & 0.81 & 0.13 & 0.81 & 0.13 & 0.82 & 0.11 & 0.87 & 0.10 & 0.93 \\
    \midrule
    \oursbf & \textbf{0.98} & N/A & \textbf{0.95} & \textbf{0.24} & \textbf{0.90} & \textbf{0.31} & \textbf{0.85} & 0.28 & \textbf{0.81} & 0.31 & \textbf{0.80} & 0.39 & \textbf{0.78} & 0.25 & \textbf{0.76} & 0.30 & \textbf{0.75} & 0.29 & \textbf{0.71} & 0.19 \\
    \bottomrule
  \end{tabular}
  \caption{Per Experience Results for CIFAR100 with 5000 Exemplars and 10 Experiences}
  \label{your-label}
\end{table*}

\begin{table*}[!ht]
  \centering
  \setlength\tabcolsep{2pt} 
  \begin{tabular}{@{}l *{20}{c}@{}}
    \toprule
    \textbf{Model} & \multicolumn{2}{c}{\textbf{Exp 1}} & \multicolumn{2}{c}{\textbf{Exp 2}} & \multicolumn{2}{c}{\textbf{Exp 3}} & \multicolumn{2}{c}{\textbf{Exp 4}} & \multicolumn{2}{c}{\textbf{Exp 5}} & \multicolumn{2}{c}{\textbf{Exp 6}} & \multicolumn{2}{c}{\textbf{Exp 7}} & \multicolumn{2}{c}{\textbf{Exp 8}} & \multicolumn{2}{c}{\textbf{Exp 9}} & \multicolumn{2}{c}{\textbf{Exp 10}} \\
    \cmidrule(lr){2-3} \cmidrule(lr){4-5} \cmidrule(lr){6-7} \cmidrule(lr){8-9} \cmidrule(lr){10-11} \cmidrule(lr){12-13} \cmidrule(lr){14-15} \cmidrule(lr){16-17} \cmidrule(lr){18-19} \cmidrule(l){20-21}
    & \textbf{FAA} & \textbf{AF} & \textbf{FAA} & \textbf{AF} & \textbf{FAA} & \textbf{AF} & \textbf{FAA} & \textbf{AF} & \textbf{FAA} & \textbf{AF} & \textbf{FAA} & \textbf{AF} & \textbf{FAA} & \textbf{AF} & \textbf{FAA} & \textbf{AF} & \textbf{FAA} & \textbf{AF} & \textbf{FAA} & \textbf{AF} \\
    \midrule
    CBM-S & 0.70 & N/A & 0.67 & 0.50 & 0.58 & 0.56 & 0.50 & 0.61 & 0.46 & 0.52 & 0.41 & 0.64 & 0.31 & 0.57 & 0.27 & 0.44 & 0.25 & 0.52 & 0.22 & 0.50 \\
    CBM-J & 0.83 & N/A & 0.64 & 0.63 & 0.52 & 0.51 & 0.48 & 0.55 & 0.45 & 0.42 & 0.36 & 0.62 & 0.33 & 0.49 & 0.25 & 0.47 & 0.27 & 0.46 & 0.22 & 0.38 \\
    ICIAP-S & 0.70 & N/A & 0.67 & 0.50 & 0.58 & 0.56 & 0.50 & 0.61 & 0.46 & 0.52 & 0.41 & 0.64 & 0.31 & 0.57 & 0.27 & 0.44 & 0.25 & 0.52 & 0.22 & 0.50 \\
    ICIAP-J & 0.83 & N/A & 0.64 & 0.63 & 0.52 & 0.51 & 0.48 & 0.55 & 0.45 & 0.42 & 0.36 & 0.62 & 0.33 & 0.49 & 0.25 & 0.47 & 0.27 & 0.46 & 0.22 & 0.38 \\
    LabelFree & 0.06 & N/A & 0.04 & 0.80 & 0.08 & 0.48 & 0.07 & 0.49 & 0.05 & 0.43 & 0.04 & 0.56 & 0.06 & 0.44 & 0.05 & 0.39 & 0.05 & 0.38 & 0.42 & 0.35 \\
    LaBo & 0.43 & N/A & 0.18 & 0.43 & 0.34 & 0.58 & 0.30 & 0.42 & 0.25 & 0.57 & 0.16 & 0.76 & 0.12 & 0.79 & 0.11 & 0.80 & 0.08 & 0.90 & 0.07 & 0.77 \\
    \midrule
    \oursbf & \textbf{0.93} & N/A & \textbf{0.84} & \textbf{0.13} & \textbf{0.84} & \textbf{0.22} & \textbf{0.81} & \textbf{0.13} & \textbf{0.81} & \textbf{0.09} & \textbf{0.79} & \textbf{0.10} & \textbf{0.78} & \textbf{0.16} & \textbf{0.76} & \textbf{0.11} & \textbf{0.75} & \textbf{0.14} & \textbf{0.76} & \textbf{0.15} \\
    \bottomrule
  \end{tabular}
  \caption{Per Experience Results for CUB (Caltech-UCSD Birds-200-2011) with 500 Exemplars and 10 Experiences}
  \label{your-label}
\end{table*}

\begin{table*}[!ht]
  \centering
  \setlength\tabcolsep{2pt} 
  \begin{tabular}{@{}l *{20}{c}@{}}
    \toprule
    \textbf{Model} & \multicolumn{2}{c}{\textbf{Exp 1}} & \multicolumn{2}{c}{\textbf{Exp 2}} & \multicolumn{2}{c}{\textbf{Exp 3}} & \multicolumn{2}{c}{\textbf{Exp 4}} & \multicolumn{2}{c}{\textbf{Exp 5}} & \multicolumn{2}{c}{\textbf{Exp 6}} & \multicolumn{2}{c}{\textbf{Exp 7}} & \multicolumn{2}{c}{\textbf{Exp 8}} & \multicolumn{2}{c}{\textbf{Exp 9}} & \multicolumn{2}{c}{\textbf{Exp 10}} \\
    \cmidrule(lr){2-3} \cmidrule(lr){4-5} \cmidrule(lr){6-7} \cmidrule(lr){8-9} \cmidrule(lr){10-11} \cmidrule(lr){12-13} \cmidrule(lr){14-15} \cmidrule(lr){16-17} \cmidrule(lr){18-19} \cmidrule(l){20-21}
    & \textbf{FAA} & \textbf{AF} & \textbf{FAA} & \textbf{AF} & \textbf{FAA} & \textbf{AF} & \textbf{FAA} & \textbf{AF} & \textbf{FAA} & \textbf{AF} & \textbf{FAA} & \textbf{AF} & \textbf{FAA} & \textbf{AF} & \textbf{FAA} & \textbf{AF} & \textbf{FAA} & \textbf{AF} & \textbf{FAA} & \textbf{AF} \\
    \midrule
    CBM-S & 0.70 & N/A & 0.68 & 0.22 & 0.63 & 0.36 & 0.58 & 0.26 & 0.58 & 0.17 & 0.55 & 0.26 & 0.48 & 0.24 & 0.46 & 0.25 & 0.45 & 0.28 & 0.46 & 0.28 \\
    CBM-J & 0.83 & N/A & 0.63 & 0.27 & 0.58 & 0.20 & 0.56 & 0.19 & 0.57 & 0.21 & 0.53 & 0.14 & 0.49 & 0.23 & 0.45 & 0.23 & 0.47 & 0.16 & 0.46 & 0.14 \\
    ICIAP-S & 0.70 & N/A & 0.68 & 0.22 & 0.63 & 0.36 & 0.58 & 0.26 & 0.58 & 0.17 & 0.55 & 0.26 & 0.48 & 0.24 & 0.46 & 0.25 & 0.45 & 0.28 & 0.46 & 0.28 \\
    ICIAP-J & 0.83 & N/A & 0.63 & 0.27 & 0.58 & 0.20 & 0.56 & 0.19 & 0.57 & 0.21 & 0.53 & 0.14 & 0.49 & 0.23 & 0.45 & 0.23 & 0.47 & 0.16 & 0.46 & 0.14 \\
    LabelFree & 0.36 & N/A & 0.19 & 0.51 & 0.27 & 0.28 & 0.00 & 0.35 & 0.00 & 0.54 & 0.00 & 0.68 & 0.00 & 0.49 & 0.00 & 0.47 & 0.00 & 0.50 & 0.58 & 0.43 \\
    LaBo & 0.44 & 0.00 & 0.20 & 0.44 & 0.36 & 0.48 & 0.31 & 0.34 & 0.27 & 0.52 & 0.20 & 0.71 & 0.13 & 0.71 & 0.13 & 0.75 & 0.09 & 0.84 & 0.07 & 0.72 \\
    \midrule
    \oursbf & \textbf{0.94} & N/A & \textbf{0.86} & \textbf{0.12} & \textbf{0.84} & \textbf{0.10} & \textbf{0.83} & \textbf{0.07} & \textbf{0.83} & \textbf{0.06} & \textbf{0.82} & \textbf{0.06} & \textbf{0.82} & \textbf{0.08} & \textbf{0.80} & \textbf{0.04} & \textbf{0.81} & \textbf{0.05} & \textbf{0.81} & \textbf{0.06} \\
    \bottomrule
  \end{tabular}
  \caption{Per Experience Results for CUB (Caltech-UCSD Birds-200-2011) with 2000 Exemplars and 10 Experiences}
  \label{your-label}
\end{table*}

\begin{table*}[!ht]
  \centering
  \setlength\tabcolsep{2pt} 
  \begin{tabular}{@{}l *{20}{c}@{}}
    \toprule
    \textbf{Model} & \multicolumn{2}{c}{\textbf{Exp 1}} & \multicolumn{2}{c}{\textbf{Exp 2}} & \multicolumn{2}{c}{\textbf{Exp 3}} & \multicolumn{2}{c}{\textbf{Exp 4}} & \multicolumn{2}{c}{\textbf{Exp 5}} & \multicolumn{2}{c}{\textbf{Exp 6}} & \multicolumn{2}{c}{\textbf{Exp 7}} & \multicolumn{2}{c}{\textbf{Exp 8}} & \multicolumn{2}{c}{\textbf{Exp 9}} & \multicolumn{2}{c}{\textbf{Exp 10}} \\
    \cmidrule(lr){2-3} \cmidrule(lr){4-5} \cmidrule(lr){6-7} \cmidrule(lr){8-9} \cmidrule(lr){10-11} \cmidrule(lr){12-13} \cmidrule(lr){14-15} \cmidrule(lr){16-17} \cmidrule(lr){18-19} \cmidrule(l){20-21}
    & \textbf{FAA} & \textbf{AF} & \textbf{FAA} & \textbf{AF} & \textbf{FAA} & \textbf{AF} & \textbf{FAA} & \textbf{AF} & \textbf{FAA} & \textbf{AF} & \textbf{FAA} & \textbf{AF} & \textbf{FAA} & \textbf{AF} & \textbf{FAA} & \textbf{AF} & \textbf{FAA} & \textbf{AF} & \textbf{FAA} & \textbf{AF} \\
    \midrule
    CBM-Seq & 0.70 & N/A & 0.68 & 0.18 & 0.63 & 0.27 & 0.58 & 0.22 & 0.59 & 0.15 & 0.56 & 0.15 & 0.51 & 0.18 & 0.49 & 0.13 & 0.49 & 0.16 & 0.52 & 0.25 \\
    CBM-J & 0.83 & N/A & 0.63 & 0.21 & 0.58 & 0.16 & 0.56 & 0.16 & 0.58 & 0.05 & 0.55 & 0.10 & 0.52 & 0.12 & 0.50 & 0.06 & 0.51 & 0.11 & 0.51 & 0.10 \\
    ICIAP-S & 0.70 & N/A & 0.68 & 0.18 & 0.63 & 0.27 & 0.58 & 0.22 & 0.59 & 0.15 & 0.56 & 0.15 & 0.51 & 0.18 & 0.49 & 0.13 & 0.49 & 0.16 & 0.52 & 0.25 \\
    ICIAP-J & 0.83 & N/A & 0.63 & 0.21 & 0.58 & 0.16 & 0.56 & 0.16 & 0.58 & 0.05 & 0.55 & 0.10 & 0.52 & 0.12 & 0.50 & 0.06 & 0.51 & 0.11 & 0.51 & 0.10 \\
    LabelFree & 0.60 & N/A & 0.15 & 0.27 & 0.17 & 0.33 & 0.25 & 0.43 & 0.24 & 0.29 & 0.11 & 0.42 & 0.16 & 0.42 & 0.11 & 0.31 & 0.07 & 0.38 & 0.62 & 0.29 \\
    LaBo & 0.44 & N/A & 0.23 & 0.43 & 0.39 & 0.54 & 0.33 & 0.29 & 0.29 & 0.47 & 0.21 & 0.73 & 0.16 & 0.66 & 0.13 & 0.77 & 0.09 & 0.8 & 0.08 & 0.64 \\
    \midrule
    \oursbf & \textbf{0.94} & N/A & \textbf{0.86} & \textbf{0.07} & \textbf{0.87} & \textbf{0.07} & \textbf{0.84} & \textbf{0.07} & \textbf{0.85} & \textbf{0.03} & \textbf{0.83} & \textbf{0.06} & \textbf{0.84} & \textbf{0.08} & \textbf{0.82} & \textbf{0.06} & \textbf{0.81} & \textbf{0.07} & \textbf{0.82} & \textbf{0.05} \\
    \bottomrule
  \end{tabular}
  \caption{Per Experience Results for CUB (Caltech-UCSD Birds-200-2011) with 5000 Exemplars and 10 Experiences}
  \label{your-label}
\end{table*}

\begin{table*}[!ht]
  \centering
  \setlength\tabcolsep{2pt} 
  \begin{tabular}{@{}l *{20}{c}@{}}
    \toprule
    \textbf{Model} & \multicolumn{2}{c}{\textbf{Exp 1}} & \multicolumn{2}{c}{\textbf{Exp 2}} & \multicolumn{2}{c}{\textbf{Exp 3}} & \multicolumn{2}{c}{\textbf{Exp 4}} & \multicolumn{2}{c}{\textbf{Exp 5}} & \multicolumn{2}{c}{\textbf{Exp 6}} & \multicolumn{2}{c}{\textbf{Exp 7}} & \multicolumn{2}{c}{\textbf{Exp 8}} & \multicolumn{2}{c}{\textbf{Exp 9}} & \multicolumn{2}{c}{\textbf{Exp 10}} \\
    \cmidrule(lr){2-3} \cmidrule(lr){4-5} \cmidrule(lr){6-7} \cmidrule(lr){8-9} \cmidrule(lr){10-11} \cmidrule(lr){12-13} \cmidrule(lr){14-15} \cmidrule(lr){16-17} \cmidrule(lr){18-19} \cmidrule(l){20-21}
    & \textbf{FAA} & \textbf{AF} & \textbf{FAA} & \textbf{AF} & \textbf{FAA} & \textbf{AF} & \textbf{FAA} & \textbf{AF} & \textbf{FAA} & \textbf{AF} & \textbf{FAA} & \textbf{AF} & \textbf{FAA} & \textbf{AF} & \textbf{FAA} & \textbf{AF} & \textbf{FAA} & \textbf{AF} & \textbf{FAA} & \textbf{AF} \\
    \midrule
    CBM-S & 0.88 & N/A & 0.85 & 0.69 & 0.69 & 0.86 & 0.58 & 0.64 & 0.46 & 0.57 & 0.40 & 0.61 & 0.29 & 0.73 & 0.29 & 0.54 & 0.26 & 0.54 & 0.21 & 0.71 \\
    CBM-J & 0.83 & N/A & 0.76 & 0.78 & 0.57 & 0.85 & 0.40 & 0.67 & 0.35 & 0.59 & 0.28 & 0.64 & 0.18 & 0.70 & 0.21 & 0.57 & 0.20 & 0.57 & 0.17 & 0.70 \\
    ICIAP-S & 0.89 & N/A & 0.59 & 0.75 & 0.55 & 0.54 & 0.44 & 0.53 & 0.35 & 0.47 & 0.29 & 0.57 & 0.22 & 0.55 & 0.20 & 0.48 & 0.17 & 0.46 & 0.14 & 0.65 \\
    ICIAP-J & 0.81 & N/A & 0.69 & 0.78 & 0.51 & 0.75 & 0.39 & 0.67 & 0.36 & 0.59 & 0.29 & 0.65 & 0.20 & 0.73 & 0.21 & 0.54 & 0.18 & 0.58 & 0.18 & 0.73 \\
    LabelFree & 0.07 & N/A & 0.07 & 0.85 & 0.09 & 0.29 & 0.04 & 0.72 & 0.10 & 0.18 & 0.05 & \textbf{0.07} & 0.10 & \textbf{0.14} & 0.09 & \textbf{0.02} & 0.04 & 0.03 & 0.11 & 0.07 \\
    LaBo & 0.56 & N/A & 0.04 & 0.55 & 0.08 & 0.50 & 0.38 & 0.56 & 0.21 & 0.52 & 0.15 & 0.54 & 0.13 & 0.68 & 0.07 & 0.62 & 0.10 & 0.7 & 0.05 & 0.80 \\
    \midrule
    \oursbf & \textbf{0.90} & N/A & \textbf{0.93} & \textbf{0.08} & \textbf{0.89} & \textbf{0.12} & \textbf{0.84} & \textbf{0.13} & \textbf{0.79} & \textbf{0.13} & \textbf{0.79} & 0.08 & \textbf{0.70} & \textbf{0.16} & \textbf{0.76} & 0.08 & \textbf{0.77} & \textbf{0.02} & \textbf{0.79} & \textbf{0.04} \\
    \bottomrule
  \end{tabular}
  \caption{Per Experience Results for ImageNet100 with 500 Exemplars and 10 Experiences}
  \label{your-label}
\end{table*}

\begin{table*}[ht]
  \centering
  \setlength\tabcolsep{2pt} 
  \begin{tabular}{@{}l *{20}{c}@{}}
    \toprule
    \textbf{Model} & \multicolumn{2}{c}{\textbf{Exp 1}} & \multicolumn{2}{c}{\textbf{Exp 2}} & \multicolumn{2}{c}{\textbf{Exp 3}} & \multicolumn{2}{c}{\textbf{Exp 4}} & \multicolumn{2}{c}{\textbf{Exp 5}} & \multicolumn{2}{c}{\textbf{Exp 6}} & \multicolumn{2}{c}{\textbf{Exp 7}} & \multicolumn{2}{c}{\textbf{Exp 8}} & \multicolumn{2}{c}{\textbf{Exp 9}} & \multicolumn{2}{c}{\textbf{Exp 10}} \\
    \cmidrule(lr){2-3} \cmidrule(lr){4-5} \cmidrule(lr){6-7} \cmidrule(lr){8-9} \cmidrule(lr){10-11} \cmidrule(lr){12-13} \cmidrule(lr){14-15} \cmidrule(lr){16-17} \cmidrule(lr){18-19} \cmidrule(l){20-21}
    & \textbf{FAA} & \textbf{AF} & \textbf{FAA} & \textbf{AF} & \textbf{FAA} & \textbf{AF} & \textbf{FAA} & \textbf{AF} & \textbf{FAA} & \textbf{AF} & \textbf{FAA} & \textbf{AF} & \textbf{FAA} & \textbf{AF} & \textbf{FAA} & \textbf{AF} & \textbf{FAA} & \textbf{AF} & \textbf{FAA} & \textbf{AF} \\
    \midrule
    CBM-S & 0.88 & N/A & 0.90 & 0.41 & 0.80 & 0.48 & 0.71 & 0.41 & 0.64 & 0.39 & 0.59 & 0.39 & 0.49 & 0.46 & 0.48 & 0.32 & 0.49 & 0.24 & 0.47 & 0.33 \\
    CBM-J & 0.83 & N/A & 0.81 & 0.51 & 0.69 & 0.50 & 0.59 & 0.43 & 0.54 & 0.36 & 0.46 & 0.45 & 0.40 & 0.57 & 0.40 & 0.38 & 0.37 & 0.31 & 0.39 & 0.42 \\
    ICIAP-S & 0.89 & N/A & 0.81 & 0.49 & 0.71 & 0.42 & 0.62 & 0.38 & 0.57 & 0.33 & 0.48 & 0.38 & 0.42 & 0.42 & 0.39 & 0.27 & 0.39 & 0.24 & 0.36 & 0.37 \\
    ICIAP-J & 0.81 & N/A & 0.78 & 0.50 & 0.67 & 0.58 & 0.57 & 0.48 & 0.51 & 0.40 & 0.47 & 0.43 & 0.40 & 0.58 & 0.39 & 0.42 & 0.36 & 0.35 & 0.37 & 0.41 \\
    LabelFree & 0.14 & N/A & 0.17 & 0.76 & 0.11 & 0.42 & 0.13 & 0.57 & 0.10 & 0.29 & 0.17 & 0.27 & 0.24 & 0.24 & 0.22 & 0.11 & 0.25 & 0.08 & 0.17 & \textbf{0.00} \\
    LaBo & 0.55 & N/A & 0.04 & 0.55 & 0.07 & 0.44 & 0.35 & 0.54 & 0.19 & 0.48 & 0.15 & 0.48 & 0.13 & 0.60 & 0.07 & 0.58 & 0.11 & 0.65 & 0.06 & 0.76 \\
    \midrule
    \oursbf & \textbf{0.90} & N/A & \textbf{0.93} & \textbf{0.07} & \textbf{0.89} & \textbf{0.09} & \textbf{0.84} & \textbf{0.11} & \textbf{0.80} & \textbf{0.09} & \textbf{0.80} & \textbf{0.07} & \textbf{0.74} & \textbf{0.16} & \textbf{0.77} & \textbf{0.07} & \textbf{0.79} & \textbf{0.01} & \textbf{0.80} & 0.04 \\
    \bottomrule
  \end{tabular}
  \caption{Per Experience Results for ImageNet100 with 2000 Exemplars and 10 Experiences}
  \label{your-label}
\end{table*}

\begin{table*}
  \centering
  \setlength\tabcolsep{2pt} 
  \begin{tabular}{@{}l *{20}{c}@{}}
    \toprule
    \textbf{Model} & \multicolumn{2}{c}{\textbf{Exp 1}} & \multicolumn{2}{c}{\textbf{Exp 2}} & \multicolumn{2}{c}{\textbf{Exp 3}} & \multicolumn{2}{c}{\textbf{Exp 4}} & \multicolumn{2}{c}{\textbf{Exp 5}} & \multicolumn{2}{c}{\textbf{Exp 6}} & \multicolumn{2}{c}{\textbf{Exp 7}} & \multicolumn{2}{c}{\textbf{Exp 8}} & \multicolumn{2}{c}{\textbf{Exp 9}} & \multicolumn{2}{c}{\textbf{Exp 10}} \\
    \cmidrule(lr){2-3} \cmidrule(lr){4-5} \cmidrule(lr){6-7} \cmidrule(lr){8-9} \cmidrule(lr){10-11} \cmidrule(lr){12-13} \cmidrule(lr){14-15} \cmidrule(lr){16-17} \cmidrule(lr){18-19} \cmidrule(l){20-21}
    & \textbf{FAA} & \textbf{AF} & \textbf{FAA} & \textbf{AF} & \textbf{FAA} & \textbf{AF} & \textbf{FAA} & \textbf{AF} & \textbf{FAA} & \textbf{AF} & \textbf{FAA} & \textbf{AF} & \textbf{FAA} & \textbf{AF} & \textbf{FAA} & \textbf{AF} & \textbf{FAA} & \textbf{AF} & \textbf{FAA} & \textbf{AF} \\
    \midrule

    CBM-S & 0.88 & N/A & 0.90 & 0.29 & 0.85 & 0.37 & 0.78 & 0.31 & 0.73 & 0.26 & 0.69 & 0.25 & 0.61 & 0.31 & 0.61 & 0.22 & 0.61 & 0.18 & 0.58 & 0.22 \\
    CBM-J & 0.83 & N/A & 0.84 & 0.35 & 0.77 & 0.34 & 0.69 & 0.37 & 0.61 & 0.32 & 0.57 & 0.39 & 0.47 & 0.42 & 0.51 & 0.33 & 0.49 & 0.21 & 0.49 & 0.33 \\
    ICIAP-S & 0.89 & N/A & 0.83 & 0.33 & 0.77 & 0.35 & 0.66 & 0.29 & 0.64 & 0.15 & 0.60 & 0.25 & 0.51 & 0.34 & 0.52 & 0.20 & 0.50 & 0.21 & 0.46 & 0.37 \\
    ICIAP-J & 0.81 & N/A & 0.83 & 0.38 & 0.75 & 0.35 & 0.68 & 0.32 & 0.60 & 0.31 & 0.55 & 0.31 & 0.48 & 0.44 & 0.50 & 0.26 & 0.49 & 0.23 & 0.49 & 0.35 \\
    LabelFree & 0.18 & N/A & 0.25 & 0.73 & 0.27 & 0.25 & 0.30 & \textbf{0.08} & 0.38 & 0.14 & 0.34 & 0.18 & 0.37 & \textbf{0.11} & 0.41 & 0.07 & 0.36 & \textbf{0.00} & 0.35 & 0.02 \\
    LaBo & 0.55 & N/A & 0.05 & 0.55 & 0.07 & 0.42 & 0.38 & 0.50 & 0.20 & 0.45 & 0.15 & 0.46 & 0.13 & 0.60 & 0.08 & 0.54 & 0.11 & 0.62 & 0.06 & 0.74 \\
    \midrule
    \oursbf & \textbf{0.90} & N/A & \textbf{0.93} & \textbf{0.07} & \textbf{0.89} & \textbf{0.12} & \textbf{0.84} & 0.11 & \textbf{0.80} & \textbf{0.09} & \textbf{0.79} & \textbf{0.09} & \textbf{0.74} & 0.16 & \textbf{0.78} & \textbf{0.05} & \textbf{0.79} & 0.03 & \textbf{0.80} & \textbf{0.02} \\
    \bottomrule
  \end{tabular}
  \caption{Per Experience Results for ImageNet100 with 5000 Exemplars and 10 Experiences}
  \label{your-label}
\end{table*}
    
\end{document}